\documentclass[11pt]{article}
\usepackage{graphicx}
\usepackage{longtable,lscape}
\usepackage{amsfonts}
\usepackage{bbm}
\usepackage{amsmath,amsthm}

\usepackage{float}
\usepackage{url}

\newcommand{\captionfonts}{\footnotesize}
\makeatletter  
\long\def\@makecaption#1#2{%
  \vskip\abovecaptionskip
  \sbox\@tempboxa{{\captionfonts #1: #2}}%
  \ifdim \wd\@tempboxa >\hsize
    {\captionfonts #1: #2\par}
  \else
    \hbox to\hsize{\hfil\box\@tempboxa\hfil}%
  \fi
  \vskip\belowcaptionskip}
\makeatother 
\begin{document}
\title{Conjunction and Negation of Natural Concepts: A Quantum-theoretic Modeling}
\author{Sandro Sozzo \\
         School of Management \\
				 University of Leicester \\
        University Road LE1 7RH Leicester, UK \\
        E-Mail: \url{ss831@le.ac.uk}} 

\maketitle
\begin{abstract}
We perform two experiments with the aim to investigate the effects of negation on the combination of natural concepts. In the first experiment, we test the membership weights of a list of exemplars with respect to two concepts, e.g., {\it Fruits} and {\it Vegetables}, and their conjunction {\it Fruits And Vegetables}. In the second experiment, we test the membership weights of the same list of exemplars with respect to the same two concepts, but negating the second, e.g., {\it Fruits} and {\it Not Vegetables}, and again their conjunction {\it Fruits And Not Vegetables}. The collected data confirm existing results on conceptual combination, namely, they show dramatic deviations from the predictions of classical (fuzzy set) logic and probability theory. More precisely, they exhibit conceptual vagueness, gradeness of membership, overextension and double overextension of membership weights with respect to the given conjunctions. Then, we show that the quantum probability model in Fock space recently elaborated to model Hampton's data on concept conjunction (Hampton, 1988a) and disjunction (Hampton, 1988b) faithfully accords with the collected data. Our quantum-theoretic modeling enables to describe these non-classical effects in terms of genuine quantum effects, namely `contextuality', `superposition', `interference' and `emergence'. The obtained results confirm and strenghten the analysis in Aerts (2009a) and Sozzo (2014) on the identification of quantum aspects in experiments on conceptual vagueness. Our results can be inserted within the general research on the identification of quantum structures in cognitive and decision processes.
\end{abstract}

\section{Introduction\label{intro}}
In the last years there has been a renewed interest in the formulation of a unified psychological theory for representing and structuring concepts. Indeed, traditional approaches to concept theory, mainly, `prototype theory' (Rosch, 1973; Rosch, 1977; Rosch, 1983), `exemplar theory' (Nosofsky, 1988; Nosofsky, 1992) and `theory theory' (Murphy \& Medin, 1985; Rumelhart \& Norman, 1988) are still facing a crucial difficulty, namely, `the problem of how modeling the combination of two or more natural concepts starting from the modeling of the component ones'. This `combination problem' has been revealed by several cognitive experiments in the last thirty years. More precisely:

(i) The `Guppy effect' in concept conjunction, also known as the `Pet-Fish problem' (Osherson \& Smith, 1981). If one measures the typicality of specific exemplars with respect to the concepts {\it Pet} and {\it Fish} and their conjunction {\it Pet-Fish}, then one experimentally finds that an exemplar such as {\it Guppy} is a very typical example of {\it Pet-Fish}, while it is neither a very typical example of {\it Pet} nor of {\it Fish}.

(ii) The deviation from classical (fuzzy) set-theoretic membership weights of exemplars with respect to pairs of concepts and their conjunction or disjunction (Hampton, 1988a,b). If one measures the membership weight of an exemplar with respect to a pair of concepts and their conjunction (disjunction), then one experimentally finds that the membership weight of the exemplar with respect to the conjunction (disjunction) is greater (less) than the membership weight of the exemplar with respect to at least one of the component concepts.

(iii) The so-called `borderline contradictions' (Alxatib \& Pelletier, 2011; Bonini, Osherson, Viale \& Williamson, 1999). Roughly speaking, a borderline contradiction is a sentence of the form $P(x) \land \lnot P(x)$, for a vague predicate $P$ and a borderline case $x$, e.g., the sentence ``Mark is rich and Mark is not rich''.

If one accepts that concepts are `graded', or `fuzzy', notions (Osherson \& Smith, 1982; Zadeh, 1965, 1982), as empirical evidence seem to confirm, then one cannot represent the membership weights and typicalities expressing such gradeness in a classical (fuzzy) set-theoretic model, where conceptual conjunctions are represented logical conjunctions and conceptual disjunctions are represented by logical disjunctions. These difficulties affect both `extensional' membership-based (Rips, 1995; Zadeh, 1982) and `intensional' attribute-based (Hampton, 1988b, 1997; Minsky, 1975). This combination problem is considered so serious that many authors maintain that not much progress is possible in the field if no light is shed on this problem (Fodor, 1994; Hampton, 1997; Kamp \& Partee, 1995; Komatsu, 1992; Rips, 1995). However no mechanism and/or procedure has as yet been identified that gives rise to a satisfactory description or explanation of the effects appearing when concepts combine.

Very similar effects and deviations from the predictions of traditional approaches have meanwhile been experienced in other domains of cognitive science, specifically, in behavioural economics (Ellsberg, 1961; Machina, 2009) and decision theory (Tversky \& Kahneman, 1983; Tversky \& Shafir, 1992). These and other difficulties have led various scholars to look for alternative approaches which could provide a more satisfactory picture of `what occurs in human thought in a cognitive or decision process'. Among the possible alternatives, a major candidate is what has been called `quantum cognition' and it rests the application of the mathematical formalism of quantum theory in cognitive and social domains (see. e.g., Aerts, 2009a,b; Aerts, Broekaert, Gabora \& Sozzo, 2013; Aerts \& Czachor, 2004;  Aerts \& Gabora, 2005a,b; Aerts, Gabora \& Sozzo, 2013; Aerts \& Sozzo, 2011, 2013; Aerts, Sozzo \& Tapia, 2014; Busemeyer \& Bruza, 2012; Busemeyer, Pothos, Franco \& Trueblood, 2011; Haven \& Khrennikov, 2013; Khrennikov, 2010; Pothos \& Busemeyer, 2009, 2013; van Rijsbergen, 2004; Wang, Busemeyer, Atmanspacher \& Pothos, 2013).

In this paper, we mainly deal with the quantum-theoretic approach to cognitive science elaborated in Brussels. This approach was motivated by a two decade research on the foundations of quantum theory (Aerts, 1999), the origins of quantum probability (Aerts, 1986; Pitowsky, 1989) and the identification of typically quantum aspects in the macroscopic world (Aerts \& Aerts, 1995; Aerts, Aerts, Broekaert \& Gabora, 2000). A {\it SCoP formalism} was worked out within the Brussels approach which relies on the interpretation of a concept as an `entity in a specific state changing under the influence of a context' rather than as a `container of instantiations' (Aerts \& Gabora, 2005a,b), and allowed the authors to provide a quantum representation of the guppy effect (Aerts \& Gabora, 2005a,b). Successively, the mathematical formalism of quantum theory was employed to model the overextension and underextension of membership weights measured by Hampton (1988a,b). More specifically, the overextension for conjunctions of concepts measured by Hampton (1988a) was described as an effect of quantum interference, superposition and emergence (Aerts, 2009a; Aerts, Gabora \& Sozzo, 2013), which also play a primary role in the description of both overextension and underextension for disjunctions of concepts (Hampton, 1988b). Furthermore, a specific conceptual combination experimentally revealed the presence of another genuine quantum effect, namely, entanglement (Aerts, 2009a,b; Aerts, Broekaert, Gabora \& Sozzo, 2012; Aerts, Gabora \& Sozzo, 2012; Aerts \& Sozzo, 2011). Finally, this quantum-theoretic framework was successfully applied to describe borderline vagueness (Sozzo, 2014).

More specifically, in the present paper we generalize Aerts (2009a)'s analysis of Hampton's overextension for the conjunction of two concepts, extending it to conjunctions and negations. Negative concepts have been typically considered as `singular concepts', since they do not have a prototype. Indeed, it is, for example, easy to determine the membership of a concept such as {\it Not Fruit}, but it does not seem that such a determination involves similarity with some prototype of {\it Not Fruit}. This is why one is naturally led to derive the negation of a concept from (fuzzy set) logical operations on the positively defined concept. There has been very little research on how human beings interpret and combine negated concepts. In this respect, Hampton (1997) performed a set of experiments in which he considered both conjunctions of the form {\it Tools Which Are Also Weapons} and conjunctions of the form {\it Tools Which Are Not Weapons}. As expected, his seminal work confirmed overextension in both conjunctions, also showing a violation of Boolean classical logical rules for the negation. These results were the starting point for our research in this paper, whose content can be summarized as follows.

In Section \ref{experiment} we describe the two experiments we performed. In the first experiment, we tested the membership weights of four different sets of exemplars with respect to four pairs $(A,B)$ of concepts and their conjunction `$A \ {\rm and} \ B$'. In the second experiment, we tested the membership weights of the same four sets of exemplars with respect to the same four pairs $(A,B)$ of concepts, but negating the second concept, hence actually considering $A$, `${\rm not} \ B$' and the conjunction `$A \ {\rm and} \ {\rm not} \ B$'. We observe that, already at this level, several exemplars exhibited overextension with respect to both `$A \ {\rm and} \ B$' and `$A \ {\rm and} \ {\rm not} \ B$', hence we get a first clue that a deviation from classical (fuzzy set) logic and probability theory is at play in our experiments. A complete analysis of the `non-classicality' underlying the collected data is presented in Section \ref{classicality} where we prove two theorems on the representability of a given set of experimental data in a classical Kolmogorovian probability space, thus extending the analysis in (Aerts, 2009a) to negated concepts. By applying these theorems, we show that a large part of our data cannot be modeled in a classical Kolmogorovian space. Moreover, we notice that the deviations from classicality are of two types: (i) overextension of membership weights with respect to both conjunctions `$A \ {\rm and} \ B$' and `$A \ {\rm and} \ {\rm not} \ B$', (ii) deviation of the negation `${\rm not} \ B$' of the concept $B$ from the classical logical negation. This non-classical behaviour led us to inquire into the possibility of representing our data in a quantum-mechanical framework. After a brief overview of the rules of a quantum-theoretic modeling in Section \ref{quantum}, we develop this modeling for the combinations `$A \ {\rm and} \ B$' and `$A \ {\rm and} \ {\rm not} \ B$' in Section \ref{brussels}, thus extending the analysis in (Aerts, 2009a). Finally, we draw our conclusions in Section \ref{model}, where we:

(i) prove that a large number of the collected data can be represented in our quantum-theoretic modeling in Fock space;

(ii) describe the observed deviations from classicality as a consequence of genuine quantum effects, such as, `contextuality', `interference', `superposition' and `emergence';

(iii) provide a further support to the explanatory hypothesis we have recently put forward for the effectiveness of a quantum approach in cognitive and decision processes. According to this hypothesis, human thought is the superposition of a `quantum emergent thought' and a `quantum logical thought', and that the quantum modeling approach applied in Fock space enables this approach to general human thought, consisting of a superposition of these two modes, to be modeled.

We observe, to conclude this section, that the results obtained in the present paper confirm those in (Aerts, 2009a) on conceptual conjunction/disjunction and in (Sozzo, 2014) on borderline vagueness. Hence they can be considered as a further theoretical support towards the identification of quantum structures in cognition.

\section{Description of the experiment\label{experiment}}
Hampton identified in his experiments systematic deviations from classical set (fuzzy set) conjunctions and disjunctions (Hampton, 1988a,b). More explicitly, if the membership weight of an exemplar $x$ with respect to the conjunction `$A \ {\rm and} \ B$' of two concepts $A$ and $B$ is higher than the membership weight of $x$ with respect to one concept (both concepts), we say that the membership weight of $x$ is `overextended' (`double overextended') with respect to the conjunction (by abuse of language, one usually says that $x$ is overextended with respect to the conjunction). If the membership weight of an exemplar $x$ with respect to the disjunction `$A \ {\rm or} \ B$' of two concepts $A$ and $B$ is less than the membership weight of $x$ with respect to one concept, we say that the membership weight of $x$ is `overextended' with respect to the disjunction (by abuse of language, one usually says that $x$ is overextended with respect to the disjunction). These were the non-classical effects detected by Hampton in the combination of two concepts. Similar effects were identified by the same author in his experiments on conjunction and negation of two concepts (Hampton, 1997). The analysis by Aerts (2009a) evidenced other deviations from classicality in Hampton's experiments. In this section we show that very similar devations from classicality can be observed in our experiment on human subjects. But we first need to describe the experiment.

In our experiment, we considered four pairs of natural concepts, namely ({\it Home Furnishing}, {\it Furniture}), ({\it Spices}, {\it Herbs}), ({\it Pets}, {\it Farmyard Animals}) and ({\it Fruits}, {\it Vegetables}). For each pair, we considered 24 exemplars and measured their membership with respect to these pairs of concepts and suitable conjunctions of these pairs. The membership was estimated by using a `7-point scale'. The tested subjects were asked to choose a number from the set $\{+3, +2, +1, 0, -1, -2, -3 \}$, where the positive numbers $+1$, $+2$ and $+3$ meant that they considered `the exemplar to be a member of the concept' -- $+3$ indicated a strong membership, $+1$ a relatively weak membership. The negative numbers $-1$, $-2$ and $-3$ meant that the subject considered `the exemplar to be a non-member of the concept' -- $-3$ indicated a strong non-membership, $-1$ a relatively weak non-membership.

For the conceptual pair ({\it Home Furnishing}, {\it Furniture}), we asked 80 subjects to estimate the membership of the first set of 24 exemplars with respect to the concepts {\it Home Furnishing}, {\it Furniture} and the negation {\it Not Furniture}. Then, we asked 40 subjects to estimate the membership of the same set of 24 exemplars with respect to the conjunctions {\it Home Furnishing And Furniture} and {\it Home Furnishing And Not Furniture}. Subsequently, we calculated the corresponding membership weights. The results are reported in Tables 1a and 1b .

For the conceptual pair ({\it Spices}, {\it Herbs}), we asked 80 subjects to estimate the membership of the second set of 24 exemplars with respect to the concepts {\it Spices}, {\it Herbs} and the negation {\it Not Herbs}. Then, we asked 40 subjects to estimate the membership of the same set of 24 exemplars with respect to the conjunctions {\it Spices And Herbs} and {\it Spices And Not Herbs}. Subsequently, we calculated the corresponding membership weights. The results are reported in Tables 2a and 2b.

For the conceptual pair ({\it Pets}, {\it Farmyard Animals}), we asked 80 subjects to estimate the membership of the third set of 24 exemplars with respect to the concepts {\it Pets}, {\it Farmyard Animals} and the negation {\it Not Farmyard Animals}. Then, we asked 40 subjects to estimate the membership of the same set of 24 exemplars with respect to the conjunctions {\it Pets And Farmyard Animals} and {\it Pets And Not Farmyard Animals}. Subsequently, we calculated the corresponding membership weights. The results are reported in Tables 3a and 3b.

For the conceptual pair ({\it Fruits}, {\it Vegetables}), we asked 80 subjects to estimate the membership of the fourth set of 24 exemplars with respect to the concepts {\it Fruits}, {\it Vegetables} and the negation {\it Not Vegetables}. Then, we asked 40 subjects to estimate the membership of the same set of 24 exemplars with respect to the conjunctions {\it Fruits And Vegetables} and {\it Fruits And Not Vegetables}. Subsequently, we calculated the corresponding membership weights. The results are reported in Tables 4a and 4b.

Pure inspection of Tables 1-4 reveals that several exemplars present overextension with respect to both conjunctions `$A  \ {\rm and} \ B$' and `$A \ {\rm and} \ {\rm not} \ B$'. For example, the membership weight of {\it Chili Pepper} with respect to {\it Spices} is 0.975, with respect to {\it Herbs} is	0.53125, while its membership weight with respect to the conjunction {\it Spices And Herbs} is 0.8 (Table 2.a), thus giving rise to overextension. Also, if we consider the membership weights of {\it Goldfish} with respect to {\it Pets} and {\it Farmward Animals}, we get 0.925	and	0.16875, respectively, while its membership weight with respect to {\it Pets And Farmyard Animals} is	0.425 (Table 3.a). Even stronger deviations in the combination {\it Fruits And Vegetables}. For example, the exemplar {\it Broccoli} scores 0.09375 with respect to {\it Fruits}, 1 with respect to {\it Vegetables}, and 0.5875 with respect to {\it Fruits And Vegetables}. A similar pattern is observed for {\it Parsley}, which scores 0.01875 with respect to {\it Fruits},	0.78125 with respect to {\it Vegetables} and 0.45 with respect to {\it Fruits And Vegetables} (Tables 4.a).

Overextension is also present when one concept is negated, that is, in the combination `$A \ {\rm and} \ {\rm not} B$'. Indeed, the membership weights of {\it Shelves} with respect to {\it Home Furnishing}, {\it Not Furniture} and {\it Home Furnishing And Not Furniture} is 0,85,	0,125 and	0.3875, respectively (Table 1.b). Then,  {\it Pepper} scores 0.99375 with respect to {\it Spices},  0.58125 with respect to {\it Not Herbs}, and	0.9 with respect to {\it Spices and Not Herbs} (Table 2.b). Finally, {\it Doberman Guard Dog} scores 0.88125 and 0.26875 with respect to {\it Pets} and {\it Farmyard Animals}, respectively, while it scores	0.55 with respect to {\it Pets And Farmyard Animals} (Table 3b).

Double overextension is also present in various cases and for both conjunctions `$A \ {\rm and} \ B$' and `$A \ {\rm and} \ {\rm not} \ B$'. For example, the membership weight of {\it Olive} with respect to {\it Fruits And Vegetables} is	0.65, which is greater than both 0.53125 and 0.63125, i.e. the membership weights of {\it Olive} with respect to {\it Fruits} and {\it Vegetables}, respectively (Table 4.a).	Furthermore, {\it Prize Bull} scores	0.13125 with respect to {\it Pets} and	0.2625 with respect to {\it Not Farmyard Animals}, but its membership weight with respect to {\it Pets And Not Farmayard Animals} is 0.275 (Table 3b).

Our preliminary analysis above already shows that manifest deviations from classicality occur in the experiment we performed. When we say `deviations from classicality' we actually mean that the collected data behave in such a way that they cannot generally be modeled by using the usual connectives of classical fuzzy set logic for conceptual conjunctions, neither the rules of classical probability for their membership weights. In order to systematically identify such deviations from classicality we need however a characterization of the representability of these data in a classical probability space. This is the content of the next section.

\bigskip
\noindent
\begin{table}[scale=0.4]
\begin{center}
\footnotesize
\begin{tabular}{|ccccccc|}
\hline
\multicolumn{7}{|l|}{\it $A$=Home Furnishing, $B$=Furniture} \\
\hline														
{\it Exemplar}		&	$\mu_{x}(A)$	&	$\mu_{x}(B)$	&	$\mu_{x}(A \ {\rm and} \ B)$	&	$\Delta_{AB}(x)$	&	$k_{AB}(x)$	&	${\rm Doub}_{AB}(x)$	\\
\hline
{\it Mantelpiece}		&	0.9	&	0.6125	&	0.7125	&	0.1	&	0.2	&	0.1875	\\
{\it Window Seat}		&	0.5	&	0.48125	&	0.45	&	-0.03125	&	0.46875	&	0.05	\\
{\it Painting}		&	0.8	&	0.4875	&	0.6375	&	0.15	&	0.35	&	0.1625	\\
{\it Light Fixture}		&	0.875	&	0.6	&	0.725	&	0.125	&	0.25	&	0.15	\\
{\it Kitchen Counter}		&	0.66875	&	0.4875	&	0.55	&	0.0625	&	0.39375	&	0.11875	\\
{\it Bath Tub}		&	0.725	&	0.5125	&	0.5875	&	0.075	&	0.35	&	0.1375	\\
{\it Deck Chair}		&	0.73125	&	0.9	&	0.7375	&	0.00625	&	0.10625	&	0.1625	\\
{\it Shelves}		&	0.85	&	0.93125	&	0.8375	&	-0.0125	&	0.05625	&	0.09375	\\
{\it Rug}		&	0.89375	&	0.575	&	0.7	&	0.125	&	0.23125	&	0.19375	\\
{\it Bed}		&	0.75625	&	0.925	&	0.7875	&	0.03125	&	0.10625	&	0.1375	\\
{\it Wall-Hangings}		&	0.86875	&	0.4625	&	0.55	&	0.0875	&	0.21875	&	0.31875	\\
{\it Space Rack}		&	0.375	&	0.425	&	0.4125	&	0.0375	&	0.6125	&	0.0125	\\
{\it Ashtray}		&	0.74375	&	0.4	&	0.4875	&	0.0875	&	0.34375	&	0.25625	\\
{\it Bar}		&	0.71875	&	0.625	&	0.6125	&	-0.0125	&	0.26875	&	0.10625	\\
{\it Lamp}		&	0.94375	&	0.64375	&	0.75	&	0.10625	&	0.1625	&	0.19375	\\
{\it Wall Mirror}		&	0.9125	&	0.75625	&	0.825	&	0.06875	&	0.15625	&	0.0875	\\
{\it Door Bell}		&	0.75	&	0.33125	&	0.5	&	0.16875	&	0.41875	&	0.25	\\
{\it Hammock}		&	0.61875	&	0.6625	&	0.6	&	-0.01875	&	0.31875	&	0.0625	\\
{\it Desk}		&	0.78125	&	0.95	&	0.775	&	-0.00625	&	0.04375	&	0.175	\\
{\it Refrigerator}		&	0.74375	&	0.725	&	0.6625	&	-0.0625	&	0.19375	&	0.08125	\\
{\it Park Bench}		&	0.53125	&	0.6625	&	0.55	&	0.01875	&	0.35625	&	0.1125	\\
{\it Waste Paper Basket}		&	0.69375	&	0.54375	&	0.5875	&	0.04375	&	0.35	&	0.10625	\\
{\it Sculpture}		&	0.825	&	0.4625	&	0.575	&	0.1125	&	0.2875	&	0.25	\\
{\it Sink Unit}		&	0.70625	&	0.56875	&	0.6	&	0.03125	&	0.325	&	0.10625	\\
\hline
\end{tabular}
\normalsize
\end{center}
{\rm Table 1a.} Membership weights with respect to the concepts {\it Home Furnishing}, {\it Furniture} and their conjunction {\it Home Furnishing And Furniture}.
\end{table}

\begin{table}[scale=0.2]
\begin{center}
\footnotesize
\begin{tabular}{|cccccccc|}
\hline
\multicolumn{8}{|l|}{\it $A$=Home Furnishing, $B$=Furniture} \\
\hline					
{\it Exemplar}		&	$\mu_{x}(A)$	&	$\mu_{x}({\rm not} \ B)$	&	$\mu_{x}(A \ {\rm and} \ {\rm not} \ B)$	&	$\Delta_{AB'}(x)$	&	$k_{AB'}(x)$	&	${\rm Doub}_{AB'}(x)$ & $l_{BB'}(x)$	\\
\hline
{\it Mantelpiece}		&	0.9	&	0.5	&	0.75	&	0.25	&	0.35	&	0.15	&	-0.1125	\\
{\it Window Seat}		&	0.5	&	0.55	&	0.4875	&	-0.0125	&	0.4375	&	0.0625	&	-0.03125	\\
{\it Painting}		&	0.8	&	0.64375	&	0.6	&	-0.04375	&	0.15625	&	0.2	&	-0.13125	\\
{\it Light Fixture}		&	0.875	&	0.5125	&	0.625	&	0.1125	&	0.2375	&	0.25	&	-0.1125	\\
{\it Kitchen Counter}		&	0.66875	&	0.61875	&	0.5375	&	-0.08125	&	0.25	&	0.13125	&	-0.10625	\\
{\it Bath Tub}		&	0.725	&	0.4625	&	0.5875	&	0.125	&	0.4	&	0.1375	&	0.025	\\
{\it Deck Chair}		&	0.73125	&	0.2	&	0.4125	&	0.2125	&	0.48125	&	0.31875	&	-0.1	\\
{\it Shelves}		&	0.85	&	0.125	&	0.3875	&	0.2625	&	0.4125	&	0.4625	&	-0.05625	\\
{\it Rug}		&	0.89375	&	0.60625	&	0.675	&	0.06875	&	0.175	&	0.21875	&	-0.18125	\\
{\it Bed}		&	0.75625	&	0.10625	&	0.3625	&	0.25625	&	0.5	&	0.39375	&	-0.03125	\\
{\it Wall-Hangings}		&	0.86875	&	0.68125	&	0.7125	&	0.03125	&	0.1625	&	0.15625	&	-0.14375	\\
{\it Space Rack}		&	0.375	&	0.61875	&	0.4875	&	0.1125	&	0.49375	&	0.13125	&	-0.04375	\\
{\it Ashtray}		&	0.74375	&	0.6375	&	0.6	&	-0.0375	&	0.21875	&	0.14375	&	-0.0375	\\
{\it Bar}		&	0.71875	&	0.50625	&	0.6125	&	0.10625	&	0.3875	&	0.10625	&	-0.13125	\\
{\it Lamp}		&	0.94375	&	0.4875	&	0.7	&	0.2125	&	0.26875	&	0.24375	&	-0.13125	\\
{\it Wall Mirror}		&	0.9125	&	0.45	&	0.6625	&	0.2125	&	0.3	&	0.25	&	-0.20625	\\
{\it Door Bell}		&	0.75	&	0.7875	&	0.6375	&	-0.1125	&	0.1	&	0.15	&	-0.11875	\\
{\it Hammock}		&	0.61875	&	0.40625	&	0.5	&	0.09375	&	0.475	&	0.11875	&	-0.06875	\\
{\it Desk}		&	0.78125	&	0.0875	&	0.325	&	0.2375	&	0.45625	&	0.45625	&	-0.0375	\\
{\it Refrigerator}		&	0.74375	&	0.40625	&	0.55	&	0.14375	&	0.4	&	0.19375	&	-0.13125	\\
{\it Park Bench}		&	0.53125	&	0.45625	&	0.2875	&	-0.16875	&	0.3	&	0.24375	&	-0.11875	\\
{\it Waste Paper Basket}		&	0.69375	&	0.63125	&	0.4125	&	-0.21875	&	0.0875	&	0.28125	&	-0.175	\\
{\it Sculpture}		&	0.825	&	0.65625	&	0.725	&	0.06875	&	0.24375	&	0.1	&	-0.11875	\\
{\it Sink Unit}		&	0.70625	&	0.575	&	0.5625	&	-0.0125	&	0.28125	&	0.14375	&	-0.14375	\\
\hline
\end{tabular}
\normalsize
\end{center}
{\rm Table 1b.} Membership weights with respect to the concepts {\it Home Furnishing}, {\it Not Furniture} and their conjunction {\it Home Furnishing And Not Furniture}.
\end{table}

\begin{table}[scale=0.2]
\begin{center}
\footnotesize
\begin{tabular}{|ccccccc|}
\hline
\multicolumn{7}{|l|}{\it $A$=Spices, $B$=Herbs} \\
\hline														
{\it Exemplar}		&	$\mu_{x}(A)$	&	$\mu_{x}(B)$	&	$\mu_{x}(A \ {\rm and} \ B)$	&	$\Delta_{AB}(x)$	&	$k_{AB}(x)$	&	${\rm Doub}_{AB}(x)$	\\
\hline
{\it Molasses}		&	0.3625	&	0.13125	&	0.2375	&	0.10625		&		0.74375	&	0.125	\\
{\it Salt}		&	0.66875	&	0.04375	&	0.2375	&	0.19375		&		0.525	&	0.43125	\\
{\it Peppermint}		&	0.66875	&	0.925	&	0.7	&	0.03125		&		0.10625	&	0.225	\\
{\it Curry}		&	0.9625	&	0.28125	&	0.5375	&	0.25625		&		0.29375	&	0.425	\\
{\it Oregano}		&	0.8125	&	0.85625	&	0.7875	&	-0.025		&		0.11875	&	0.06875	\\
{\it MSG}		&	0.44375	&	0.11875	&	0.225	&	0.10625		&		0.6625	&	0.21875	\\
{\it Chili Pepper}		&	0.975	&	0.53125	&	0.8	&	0.26875		&		0.29375	&	0.175	\\
{\it Mustard}		&	0.65	&	0.275	&	0.4875	&	0.2125		&		0.5625	&	0.1625	\\
{\it Mint}		&	0.64375	&	0.95625	&	0.7875	&	0.14375		&		0.1875	&	0.16875	\\
{\it Cinnamon}		&	1	&	0.49375	&	0.6875	&	0.19375		&		0.19375	&	0.3125	\\
{\it Parsley}		&	0.5375	&	0.9	&	0.675	&	0.1375		&		0.2375	&	0.225	\\
{\it Saccarin}		&	0.34375	&	0.1375	&	0.2375	&	0.1		&		0.75625	&	0.10625	\\
{\it Poppy Seeds}		&	0.81875	&	0.46875	&	0.5875	&	0.11875		&		0.3	&	0.23125	\\
{\it Pepper}		&	0.99375	&	0.46875	&	0.7	&	0.23125		&		0.2375	&	0.29375	\\
{\it Turmeric}		&	0.88125	&	0.525	&	0.7375	&	0.2125		&		0.33125	&	0.14375	\\
{\it Sugar}		&	0.45	&	0.34375	&	0.35	&	0.00625		&		0.55625	&	0.1	\\
{\it Vinegar}		&	0.3	&	0.10625	&	0.15	&	0.04375		&		0.74375	&	0.15	\\
{\it Sesame Seeds}		&	0.8	&	0.4875	&	0.5875	&	0.1		&		0.3	&	0.2125	\\
{\it Lemon Juice}		&	0.275	&	0.2	&	0.15	&	-0.05		&		0.675	&	0.125	\\
{\it Chocolate}		&	0.26875	&	0.2125	&	0.2	&	-0.0125		&		0.71875	&	0.06875	\\
{\it Horseradish}		&	0.6125	&	0.66875	&	0.6125	&	0		&		0.33125	&	0.05625	\\
{\it Vanilla}		&	0.7625	&	0.5125	&	0.625	&	0.1125		&		0.35	&	0.1375	\\
{\it Chives}		&	0.6625	&	0.8875	&	0.7625	&	0.1		&		0.2125	&	0.125	\\
{\it Root Ginger}		&	0.84375	&	0.5625	&	0.6875	&	0.125		&		0.28125	&	0.15625	\\
\hline
\end{tabular}
\normalsize
\end{center}
{\rm Table 2a.} Membership weights with respect to the concepts {\it Spices}, {\it Herbs} and their conjunction {\it Spices And Herbs}.
\end{table}

\begin{table}[scale=0.2]
\begin{center}
\footnotesize
\begin{tabular}{|cccccccc|}
\hline
\multicolumn{8}{|l|}{\it $A$=Spices, $B$=Herbs} \\
\hline					
{\it Exemplar}		&	$\mu_{x}(A)$	&	$\mu_{x}({\rm not} \ B)$	&	$\mu_{x}(A \ {\rm and} \ {\rm not} \ B)$	&	$\Delta_{AB'}(x)$	&	$k_{AB'}(x)$	&	${\rm Doub}_{AB'}(x)$ & $l_{BB'}(x)$	\\
\hline
{\it Molasses}		&	0.3625	&	0.8375	&	0.5375	&	0.175	&	0.3375	&	0.3	&	0.03125	\\
{\it Salt}		&	0.66875	&	0.91875	&	0.6875	&	0.01875	&	0.1	&	0.23125	&	0.0375	\\
{\it Peppermint}		&	0.66875	&	0.1	&	0.375	&	0.275	&	0.60625	&	0.29375	&	-0.025	\\
{\it Curry}		&	0.9625	&	0.775	&	0.875	&	0.1	&	0.1375	&	0.0875	&	-0.05625	\\
{\it Oregano}		&	0.8125	&	0.125	&	0.4	&	0.275	&	0.4625	&	0.4125	&	0.01875	\\
{\it MSG}		&	0.44375	&	0.85	&	0.575	&	0.13125	&	0.28125	&	0.275	&	0.03125	\\
{\it Chili Pepper}		&	0.975	&	0.5625	&	0.9	&	0.3375	&	0.3625	&	0.075	&	-0.09375	\\
{\it Mustard}		&	0.65	&	0.70625	&	0.65	&	0	&	0.29375	&	0.05625	&	0.01875	\\
{\it Mint}		&	0.64375	&	0.0875	&	0.3125	&	0.225	&	0.58125	&	0.33125	&	-0.04375	\\
{\it Cinnamon}		&	1	&	0.5125	&	0.7875	&	0.275	&	0.275	&	0.2125	&	-0.00625	\\
{\it Parsley}		&	0.5375	&	0.0875	&	0.2625	&	0.175	&	0.6375	&	0.275	&	0.0125	\\
{\it Saccarin}		&	0.34375	&	0.875	&	0.5375	&	0.19375	&	0.31875	&	0.3375	&	-0.0125	\\
{\it Poppy Seeds}		&	0.81875	&	0.5375	&	0.6625	&	0.125	&	0.30625	&	0.15625	&	-0.00625	\\
{\it Pepper}		&	0.99375	&	0.58125	&	0.9	&	0.31875	&	0.325	&	0.09375	&	-0.05	\\
{\it Turmeric}		&	0.88125	&	0.43125	&	0.6875	&	0.25625	&	0.375	&	0.19375	&	0.04375	\\
{\it Sugar}		&	0.45	&	0.76875	&	0.5625	&	0.1125	&	0.34375	&	0.20625	&	-0.1125	\\
{\it Vinegar}		&	0.3	&	0.88125	&	0.4125	&	0.1125	&	0.23125	&	0.46875	&	0.0125	\\
{\it Sesame Seeds}		&	0.8	&	0.5875	&	0.7	&	0.1125	&	0.3125	&	0.1	&	-0.075	\\
{\it Lemon Juice}		&	0.275	&	0.80625	&	0.425	&	0.15	&	0.34375	&	0.38125	&	-0.00625	\\
{\it Chocolate}		&	0.26875	&	0.8	&	0.4625	&	0.19375	&	0.39375	&	0.3375	&	-0.0125	\\
{\it Horseradish}		&	0.6125	&	0.28125	&	0.4	&	0.11875	&	0.50625	&	0.2125	&	0.05	\\
{\it Vanilla}		&	0.7625	&	0.4875	&	0.6125	&	0.125	&	0.3625	&	0.15	&	0	\\
{\it Chives}		&	0.6625	&	0.25625	&	0.275	&	0.01875	&	0.35625	&	0.3875	&	-0.14375	\\
{\it Root Ginger}		&	0.84375	&	0.44375	&	0.5875	&	0.14375	&	0.3	&	0.25625	&	-0.00625	\\
\hline
\end{tabular}
\normalsize
\end{center}
{\rm Table 2b.} Membership weights with respect to the concepts {\it Spices}, {\it Not Herbs} and their conjunction {\it Spices And Not Herbs}.
\end{table}

\begin{table}[scale=0.2]
\begin{center}
\footnotesize
\begin{tabular}{|ccccccc|}
\hline
\multicolumn{7}{|l|}{\it $A$=Pets, $B$=Farmayard Animals} \\
\hline														
{\it Exemplar}		&	$\mu_{x}(A)$	&	$\mu_{x}(B)$	&	$\mu_{x}(A \ {\rm and} \ B)$	&	$\Delta_{AB}(x)$	&	$k_{AB}(x)$	&	${\rm Doub}_{AB}(x)$	\\
\hline
{\it Goldfish}		&	0.925	&	0.16875	&	0.425	&	0.25625	&	0.33125	&	0.5	\\
{\it Robin}		&	0.275	&	0.3625	&	0.3125	&	0.0375	&	0.675	&	0.05	\\
{\it Blue-tit}		&	0.25	&	0.3125	&	0.175	&	-0.075	&	0.6125	&	0.1375	\\
{\it Collie Dog}		&	0.95	&	0.76875	&	0.8625	&	0.09375	&	0.14375	&	0.0875	\\
{\it Camel}		&	0.15625	&	0.25625	&	0.2	&	0.04375	&	0.7875	&	0.05625	\\
{\it Squirrel}		&	0.3	&	0.39375	&	0.275	&	-0.025	&	0.58125	&	0.11875	\\
{\it Guide Dog for Blind}		&	0.925	&	0.325	&	0.55	&	0.225	&	0.3	&	0.375	\\
{\it Spider}		&	0.3125	&	0.3875	&	0.3125	&	0	&	0.6125	&	0.075	\\
{\it Homing Pigeon}		&	0.40625	&	0.70625	&	0.5625	&	0.15625	&	0.45	&	0.14375	\\
{\it Monkey}		&	0.39375	&	0.175	&	0.2	&	0.025	&	0.63125	&	0.19375	\\
{\it Circus Horse}		&	0.3	&	0.48125	&	0.3375	&	0.0375	&	0.55625	&	0.14375	\\
{\it Prize Bull}		&	0.13125	&	0.7625	&	0.425	&	0.29375	&	0.53125	&	0.3375	\\
{\it Rat}		&	0.2	&	0.35625	&	0.2125	&	0.0125	&	0.65625	&	0.14375	\\
{\it Badger}		&	0.1625	&	0.275	&	0.1375	&	-0.025	&	0.7	&	0.1375	\\
{\it Siamese Cat}		&	0.9875	&	0.5	&	0.7375	&	0.2375	&	0.25	&	0.25	\\
{\it Race Horse}		&	0.2875	&	0.7	&	0.5125	&	0.225	&	0.525	&	0.1875	\\
{\it Fox}		&	0.13125	&	0.3	&	0.175	&	0.04375	&	0.74375	&	0.125	\\
{\it Donkey}		&	0.2875	&	0.9	&	0.5625	&	0.275	&	0.375	&	0.3375	\\
{\it Field Mouse}		&	0.1625	&	0.40625	&	0.225	&	0.0625	&	0.65625	&	0.18125	\\
{\it Ginger Tom-cat}		&	0.81875	&	0.50625	&	0.5875	&	0.08125	&	0.2625	&	0.23125	\\
{\it Husky in Slead Team}		&	0.64375	&	0.50625	&	0.5625	&	0.05625	&	0.4125	&	0.08125	\\
{\it Cart Horse}		&	0.26875	&	0.8625	&	0.525	&	0.25625	&	0.39375	&	0.3375	\\
{\it Chicken}		&	0.23125	&	0.95	&	0.575	&	0.34375	&	0.39375	&	0.375	\\
{\it Doberman Guard Dog}		&	0.88125	&	0.75625	&	0.8	&	0.04375	&	0.1625	&	0.08125	\\
\hline
\end{tabular}
\normalsize
\end{center}
{\rm Table 3a.} Membership weights with respect to the concepts {\it Pets}, {\it Farmyard Animals} and their conjunction {\it Pets And Farmayard Animals}.
\end{table}

\begin{table}[scale=0.2]
\begin{center}
\footnotesize
\begin{tabular}{|cccccccc|}
\hline
\multicolumn{8}{|l|}{\it $A$=Pets, $B$=Farmyard Animals} \\
\hline					
{\it Exemplar}		&	$\mu_{x}(A)$	&	$\mu_{x}({\rm not} \ B)$	&	$\mu_{x}(A \ {\rm and} \ {\rm not} \ B)$	&	$\Delta_{AB'}(x)$	&	$k_{AB'}(x)$	&	${\rm Doub}_{AB'}(x)$ & $l_{BB'}(x)$	\\
\hline
{\it Goldfish}		&	0.925	&	0.8125	&	0.9125	&	0.1	&	0.175	&	0.0125	&	0.01875	\\
{\it Robin}		&	0.275	&	0.6375	&	0.35	&	0.075	&	0.4375	&	0.2875	&	0	\\
{\it Blue-tit}		&	0.25	&	0.7125	&	0.3875	&	0.1375	&	0.425	&	0.325	&	-0.025	\\
{\it Collie Dog}		&	0.95	&	0.35	&	0.5625	&	0.2125	&	0.2625	&	0.3875	&	-0.11875	\\
{\it Camel}		&	0.15625	&	0.75	&	0.3125	&	0.15625	&	0.40625	&	0.4375	&	-0.00625	\\
{\it Squirrel}		&	0.3	&	0.65	&	0.2625	&	-0.0375	&	0.3125	&	0.3875	&	-0.04375	\\
{\it Guide Dog for Blind}		&	0.925	&	0.69375	&	0.725	&	0.03125	&	0.10625	&	0.2	&	-0.01875	\\
{\it Spider}		&	0.3125	&	0.63125	&	0.3125	&	0	&	0.36875	&	0.31875	&	-0.01875	\\
{\it Homing Pigeon}		&	0.40625	&	0.3375	&	0.25	&	-0.0875	&	0.50625	&	0.15625	&	-0.04375	\\
{\it Monkey}		&	0.39375	&	0.79375	&	0.4875	&	0.09375	&	0.3	&	0.30625	&	0.03125	\\
{\it Circus Horse}		&	0.3	&	0.6	&	0.35	&	0.05	&	0.45	&	0.25	&	-0.08125	\\
{\it Prize Bull}		&	0.13125	&	0.2625	&	0.275	&	0.14375	&	0.88125	&	-0.0125	&	-0.025	\\
{\it Rat}		&	0.2	&	0.675	&	0.275	&	0.075	&	0.4	&	0.4	&	-0.03125	\\
{\it Badger}		&	0.1625	&	0.73125	&	0.2625	&	0.1	&	0.36875	&	0.46875	&	-0.00625	\\
{\it Siamese Cat}		&	0.9875	&	0.525	&	0.75	&	0.225	&	0.2375	&	0.2375	&	-0.025	\\
{\it Race Horse}	&	0.2875	&	0.3875	&	0.3125	&	0.025	&	0.6375	&	0.075	&	-0.0875	\\
{\it Fox}		&	0.13125	&	0.68125	&	0.2875	&	0.15625	&	0.475	&	0.39375	&	0.01875	\\
{\it Donkey}		&	0.2875	&	0.15	&	0.175	&	0.025	&	0.7375	&	0.1125	&	-0.05	\\
{\it Field Mouse}		&	0.1625	&	0.5875	&	0.2375	&	0.075	&	0.4875	&	0.35	&	0.00625	\\
{\it Ginger Tom-cat}		&	0.81875	&	0.54375	&	0.575	&	0.03125	&	0.2125	&	0.24375	&	-0.05	\\
{\it Husky in Slead team}		&	0.64375	&	0.525	&	0.5125	&	-0.0125	&	0.34375	&	0.13125	&	-0.03125	\\
{\it Cart Horse}		&	0.26875	&	0.15	&	0.2	&	0.05	&	0.78125	&	0.06875	&	-0.0125	\\
{\it Chicken}		&	0.23125	&	0.0625	&	0.1125	&	0.05	&	0.81875	&	0.11875	&	-0.0125	\\
{\it Doberman Guard Dog}		&	0.88125	&	0.26875	&	0.55	&	0.28125	&	0.4	&	0.33125	&	-0.025	\\
\hline
\end{tabular}
\normalsize
\end{center}
{\rm Table 3b.} Membership weights with respect to the concepts {\it Pets}, {\it Not Farmyard Animals} and their conjunction {\it Pets And Not Farmyard Animals}.
\end{table}

\begin{table}[scale=0.2]
\begin{center}
\footnotesize
\begin{tabular}{|ccccccc|}
\hline
\multicolumn{7}{|l|}{\it $A$=Fruits, $B$=Vegetables} \\
\hline														
{\it Exemplar}		&	$\mu_{x}(A)$	&	$\mu_{x}(B)$	&	$\mu_{x}(A \ {\rm and} \ B)$	&	$\Delta_{AB}(x)$	&	$k_{AB}(x)$	&	${\rm Doub}_{AB}(x)$	\\
\hline
{\it Apple}		&	1	&	0.225	&	0.6	&	0.375	&	0.375	&	0.4	\\
{\it Parsley}		&	0.01875	&	0.78125	&	0.45	&	0.43125	&	0.65	&	0.33125	\\
{\it Olive}		&	0.53125	&	0.63125	&	0.65	&	0.11875	&	0.4875	&	-0.01875	\\
{\it Chili Pepper}		&	0.1875	&	0.73125	&	0.5125	&	0.325	&	0.59375	&	0.21875	\\
{\it Broccoli}		&	0.09375	&	1	&	0.5875	&	0.49375	&	0.49375	&	0.4125	\\
{\it Root Ginger}		&	0.1375	&	0.7125	&	0.4625	&	0.325	&	0.6125	&	0.25	\\
{\it Pumpkin}		&	0.45	&	0.775	&	0.6625	&	0.2125	&	0.4375	&	0.1125	\\
{\it Raisin}		&	0.88125	&	0.26875	&	0.525	&	0.25625	&	0.375	&	0.35625	\\
{\it Acorn}		&	0.5875	&	0.4	&	0.4625	&	0.0625	&	0.475	&	0.125	\\
{\it Mustard}		&	0.06875	&	0.3875	&	0.2875	&	0.21875	&	0.83125	&	0.1	\\
{\it Rice}		&	0.11875	&	0.45625	&	0.2125	&	0.09375	&	0.6375	&	0.24375	\\
{\it Tomato}		&	0.3375	&	0.8875	&	0.7	&	0.3625	&	0.475	&	0.1875	\\
{\it Coconut}		&	0.925	&	0.31875	&	0.5625	&	0.24375	&	0.31875	&	0.3625	\\
{\it Mushroom}		&	0.11875	&	0.6625	&	0.325	&	0.20625	&	0.54375	&	0.3375	\\
{\it Wheat}		&	0.16875	&	0.50625	&	0.3375	&	0.16875	&	0.6625	&	0.16875	\\
{\it Green Pepper}		&	0.225	&	0.6125	&	0.4875	&	0.2625	&	0.65	&	0.125	\\
{\it Watercress}		&	0.1375	&	0.7625	&	0.4875	&	0.35	&	0.5875	&	0.275	\\
{\it Peanut}		&	0.61875	&	0.29375	&	0.475	&	0.18125	&	0.5625	&	0.14375	\\
{\it Black Pepper}		&	0.20625	&	0.4125	&	0.375	&	0.16875	&	0.75625	&	0.0375	\\
{\it Garlic}		&	0.125	&	0.7875	&	0.525	&	0.4	&	0.6125	&	0.2625	\\
{\it Yam}		&	0.375	&	0.65625	&	0.5875	&	0.2125	&	0.55625	&	0.06875	\\
{\it Elderberry}		&	0.50625	&	0.39375	&	0.45	&	0.05625	&	0.55	&	0.05625	\\
{\it Almond}		&	0.7625	&	0.29375	&	0.475	&	0.18125	&	0.41875	&	0.2875	\\
{\it Lentils}		&	0.1125	&	0.6625	&	0.375	&	0.2625	&	0.6	&	0.2875	\\
\hline
\end{tabular}
\normalsize
\end{center}
{\rm Table 4a.} Membership weights with respect to the concepts {\it Fruits}, {\it Vegetables} and their conjunction {\it Fruits And Vegetables}.
\end{table}

\begin{table}[scale=0.2]
\begin{center}
\footnotesize
\begin{tabular}{|cccccccc|}
\hline
\multicolumn{8}{|l|}{\it $A$=Fruits, $B$=Vegetables} \\
\hline					
{\it Exemplar}		&	$\mu_{x}(A)$	&	$\mu_{x}({\rm not} \ B)$	&	$\mu_{x}(A \ {\rm and} \ {\rm not} \ B)$	&	$\Delta_{AB'}(x)$	&	$k_{AB'}(x)$	&	${\rm Doub}_{AB'}(x)$ & $l_{BB'}(x)$	\\
\hline
{\it Apple}		&	1	&	0.81875	&	0.8875	&	0.06875	&	0.06875	&	0.1125	&		-0.04375 \\
{\it Parsley}		&	0.01875	&	0.25	&	0.1	&	0.08125	&	0.83125	&	0.15	&		-0.03125 \\
{\it Olive}		&	0.53125	&	0.44375	&	0.3375	&	-0.10625	&	0.3625	&	0.19375	&		-0.075 \\
{\it Chili Pepper}		&	0.1875	&	0.35	&	0.2	&	0.0125	&	0.6625	&	0.15	&		-0.08125 \\
{\it Broccoli}		&	0.09375	&	0.0625	&	0.0875	&	0.025	&	0.93125	&	0.00625	&		-0.0625 \\
{\it Root Ginger}		&	0.1375	&	0.325	&	0.1375	&	0	&	0.675	&	0.1875	&		-0.0375 \\
{\it Pumpkin}		&	0.45	&	0.2625	&	0.2125	&	-0.05	&	0.5	&	0.2375	&		-0.0375 \\
{\it Raisin}		&	0.88125	&	0.7625	&	0.75	&	-0.0125	&	0.10625	&	0.13125	&		-0.03125 \\
{\it Acorn}		&	0.5875	&	0.64375	&	0.4875	&	-0.1	&	0.25625	&	0.15625	&		-0.04375 \\
{\it Mustard}		&	0.06875	&	0.6	&	0.225	&	0.15625	&	0.55625	&	0.375	&		0.0125  \\
{\it Rice}		&	0.11875	&	0.51875	&	0.225	&	0.10625	&	0.5875	&	0.29375	&		0.025 \\
{\it Tomato}		&	0.3375	&	0.1875	&	0.2	&	0.0125	&	0.675	&	0.1375	&		-0.075  \\
{\it Coconut}		&	0.925	&	0.7	&	0.6875	&	-0.0125	&	0.0625	&	0.2375	&		-0.01875 \\
{\it Mushroom}		&	0.11875	&	0.38125	&	0.125	&	0.00625	&	0.625	&	0.25625	&		-0.04375 \\
{\it Wheat}		&	0.16875	&	0.51875	&	0.2125	&	0.04375	&	0.525	&	0.30625	&		-0.025 \\
{\it Green Pepper}		&	0.225	&	0.40625	&	0.2375	&	0.0125	&	0.60625	&	0.16875	&		-0.01875 \\
{\it Watercress}		&	0.1375	&	0.25	&	0.1	&	-0.0375	&	0.7125	&	0.15	&		-0.0125 \\
{\it Peanut}		&	0.61875	&	0.75	&	0.55	&	-0.06875	&	0.18125	&	0.2	&		-0.04375 \\
{\it Black Pepper}		&	0.20625	&	0.6125	&	0.2125	&	0.00625	&	0.39375	&	0.4	&		-0.025 \\
{\it Garlic}		&	0.125	&	0.24375	&	0.1	&	-0.025	&	0.73125	&	0.14375	&		-0.03125 \\
{\it Yam}		&	0.375	&	0.43125	&	0.2375	&	-0.1375	&	0.43125	&	0.19375	&		-0.0875 \\
{\it Elderberry}		&	0.50625	&	0.60625	&	0.4125	&	-0.09375	&	0.3	&	0.19375	&		0 \\
{\it Almond}		&	0.7625	&	0.71875	&	0.6125	&	-0.10625	&	0.13125	&	0.15	&		-0.0125 \\
{\it Lentils}		&	0.1125	&	0.375	&	0.1125	&	0	&	0.625	&	0.2625	&		-0.0375 \\
\hline
\end{tabular}
\normalsize
\end{center}
{\rm Table 4b.} Membership weights with respect to the concepts {\it Fruits}, {\it Not Vegetables} and their conjunction {\it Pets And Not Vegetables}.
\end{table}

\section{Classical models for conjunctions and negations of two concepts\label{classicality}}
We derive in this section necessary and sufficient conditions for the classicality of experimental data coming from the membership weights of two concepts $A$ and $B$ with respect to the conceptual negation `${\rm not} \ B$' and the conjunctions `$A \ \rm{and} \ B$' and `$A \ \rm{and} \ {\rm not} \ B$'. More explicitly, we first derive the constraints that should be satisfied by the membership weights $\mu_{x}(A)$, $\mu_{x}(B)$ and $\mu_{x}(A \ {\rm and} \ B)$ of the exemplar $x$ with respect to the concepts $A$, $B$ and `$A \ \rm{and} \ B$', respectively, in order to represent these data in a classical probability model satisfying the axioms of Kolmogorov. Then, we derive the constraints that should be satisfied by the membership weights $\mu_{x}(A)$, $\mu_{x}({\rm not } \ B)$ and $\mu_{x}(A \ {\rm and} \ {\rm not } \ B)$ of the exemplar $x$ with respect to the concepts $A$, $B$,`${\rm not} \ B$' and `$A \ \rm{and} \ {\rm not} \ B$', respectively, in order to represent these data in a classical Kolmogorovian probability model. We follow here mathematical procedures that are similar to those employed in Aerts (2009a) for the classicality of conceptual conjunctions and disjunctions. Let us start by clearly defining what we mean by the notion of `classical', or `Kolmogorovian', probability model.

Let us start by the definition of a $\sigma$-algebra over a set.

\bigskip
\noindent
\textbf{Definition 1.}
{\it A \emph{$\sigma$-algebra} over a set $\Omega$ is a non-empty collection $\sigma(\Omega)$ of subsets of $\Omega$ that is closed under complementation and countable unions of its members. It is a Boolean algebra, completed to include countably infinite operations.}

\bigskip
\noindent
Measure structures are the most general classical structures devised by mathematicians and physicists to structure weights. A Kolmogorovian probability measure is such a measure applied to statistical data. It is called `Kolmogorovian', because Andrey Kolmogorov was the first to axiomatize probability theory in this way (Kolmogorov, 1933).

\bigskip
\noindent
\textbf{Definition 2.}
{\it A \emph{measure} $P$ is a function defined on a $\sigma$-algebra $\sigma(\Omega)$ over a set $\Omega$ and taking values in the extended interval $[0,\infty]$ such that the following three conditions are satisfied:

(i) the empty set has measure zero;

(ii) if  $E_1$, $E_2$, $E_3$, $\dots$ is a countable sequence of pairwise disjoint sets in $\sigma(\Omega)$, the measure of the union of all the $E_i$ is equal to the sum of the measures of each $E_i$ \emph{(countable additivity}, or \emph{$\sigma$-additivity)};

(iii) the triple $(\Omega,\sigma(\Omega),P)$ satisfying (i) and (ii) is then called a measure space, and the members of $\sigma(\Omega)$ are called measurable sets.

A \emph{Kolmogorovian probability measure} is a measure with total measure one. A Kolmogorovian probability space $(\Omega,\sigma(\Omega),P)$ is a measure space $(\Omega,\sigma(\Omega),P)$ such that $P$ is a Kolmogorovian probability. The three conditions expressed in a mathematical way are:
\begin{equation} \label{KolmogorovianMeasure}
P(\emptyset)=0 \quad P(\bigcup_{i=1}^\infty E_i)=\sum_{i=1}^\infty P(E_i) \quad P(\Omega)=1
\end{equation}
}

\bigskip
\noindent
Let us now come to the possibility to represent a set of experimental data on two concepts and their conjunction in a classical Kolmogorovian probability model.

\bigskip
\noindent
\textbf{Definition 3.}
{\it We say that the membership weights $\mu_{x}(A)$, $\mu_{x}(B)$ and $\mu_{x}(A\ {\rm and}\ B)$ of the exemplar $x$ with respect to the pair of concepts $A$ and $B$ and their conjunction `$A$ and $B$', respectively, can be represented in a \emph{classical Kolmogorovian probability model} if there exists a Kolmogorovian probability space $(\Omega,\sigma(\Omega),P)$ and events $E_A, E_B \in \sigma(\Omega)$ of the events algebra $\sigma(\Omega)$ such that
\begin{equation}
P(E_A) = \mu_{x}(A) \quad P(E_B) = \mu_{x}(B) \quad {\rm and} \quad P(E_A \cap E_B) = \mu_{x}(A\ {\rm and}\ B)
\end{equation}}
We can prove a useful theorem on the representability of the membership weights with respect to two concepts and their conjunction.

\bigskip
\noindent
\textbf{Theorem 1.} {\it The membership weights $\mu_{x}(A), \mu_{x}(B)$ and $\mu_{x}(A\ {\rm and}\ B)$ of the exemplar $x$ with respect to concepts $A$ and $B$ and their conjunction `$A$ and $B$', respectively, can be represented in a classical Kolmogorovian probability model if and only if they satisfy the following inequalities:}
\begin{eqnarray} \label{ineq01}
&0 \le \mu_{x}(A\ {\rm and}\ B) \le \mu_{x}(A) \le 1 \\ \label{ineq02}
&0 \le \mu_{x}(A\ {\rm and}\ B) \le \mu_{x}(B) \le 1 \\ \label{ineq03}
& \mu_{x}(A) + \mu_{x}(B) - \mu_{x}(A\ {\rm and}\ B) \le 1
\end{eqnarray}
\textbf{Proof.}
If $\mu_{x}(A), \mu_{x}(B)$ and $\mu_{x}(A\ {\rm and}\ B)$ can be represented in a classical probability model, then there exists a Kolmogorovian probability space $(\Omega,\sigma(\Omega),P)$ and events $E_A, E_B \in \sigma(\Omega)$ such that $P(E_A) = \mu_{x}(A)$, $P(E_B) = \mu_{x}(B)$ and $P(E_A \cap E_B) = \mu_{x}(A\ {\rm and}\ B)$. From the general properties of a Kolmogorovian probability space it follows that we have $0 \le P(E_A \cap E_B) \le P(E_A) \le 1$ and $0 \le P(E_A \cap E_B) \le P(E_B) \le 1$, which proves that inequalities (\ref{ineq01}) and (\ref{ineq02}) are satisfied. From the same general properties of a Kolmogorovian probability space it also follows that we have $P(E_A \cup E_B) = P(E_A) + P(E_B) - P(E_A \cap E_B)$, and since $P(E_A \cup E_B) \le 1$ we also have $P(E_A) + P(E_B) - P(E_A \cap E_B) \le 1$. This proves that inequality (\ref{ineq03}) is satisfied. We have now proved that for the classical conjunction data $\mu_{x}(A), \mu_{x}(B)$ and $\mu_{x}(A\ {\rm and}\ B)$ the three inequalities are satisfied.

Now suppose that we have an exemplar $x$ whose membership weights $\mu_{x}(A), \mu_{x}(B), \mu_{x}(A\ {\rm and}\ B)$ with respect to the concepts $A$ and $B$ and their conjunction `$A$ and $B$' are such that inequalities (\ref{ineq01}), (\ref{ineq02}) and (\ref{ineq03}) are satisfied. We prove that, as a consequence, $\mu_{x}(A), \mu_{x}(B)$ and $\mu_{x}(A\ {\rm and}\ B)$ can be represented in a Kolmogorovian probability model. To this end we explicitly construct a Kolmogorovian probability space that models these data.
Consider the set $\Omega=\{1, 2, 3, 4\}$ and $\sigma(\Omega) = {\cal P}(\Omega)$, the set of all subsets of $\Omega$. We define
\begin{eqnarray} \label{eqtheorem101}
&P(\{1\}) = \mu_{x}(A\ {\rm and}\ B) \\ \label{eqtheorem102}
&P(\{2\}) = \mu_{x}(A) - \mu_{x}(A\ {\rm and}\ B) \\ \label{eqtheorem103}
&P(\{3\}) = \mu_{x}(B) - \mu_{x}(A\ {\rm and}\ B) \\ \label{eqtheorem104}
&P(\{4\}) = 1-\mu_{x}(A)-\mu_{x}(B)+\mu_{x}(A\ {\rm and}\ B)
\end{eqnarray}
and further for an arbitrary subset $S \subseteq \{1,2,3,4\}$ we define
\begin{equation} \label{defarbitrarysubset}
P(S) = \sum_{a\in S}P(\{a\})
\end{equation}
Let us prove that $P: \sigma(\Omega) \rightarrow [0,1]$ is a probability measure. To this end we need to prove that $P(S) \in [0,1]$ for an arbitrary subset $S \subseteq \Omega$, and that the `sum formula' for a probability measure is satisfied to comply with (\ref{KolmogorovianMeasure}). The sum formula for a probability measure is satisfied because of definition (\ref{defarbitrarysubset}). What remains to be proved is that $P(S) \in [0,1]$ for an arbitrary subset $S \subseteq \Omega$. $P(\{1\}), P(\{2\}), P(\{3\})$ and $P(\{4\})$ are contained in $[0,1]$ as a direct consequence of inequalities (\ref{ineq01}), (\ref{ineq02}) and (\ref{ineq03}). Further, we have $P(\{1,2\}) = \mu_{x}(A), P(\{1,3\})=\mu_{x}(B), P(\{3,4\}) = 1-\mu_{x}(A), P(\{2,4\}) = 1-\mu_{x}(B), P(\{2,3,4\}) = 1-\mu_{x}(A\ {\rm and}\ B)$ and $P(\{1,2,3\}) = \mu_{x}(A)+\mu_{x}(B)-\mu_{x}(A\ {\rm and}\ B)$, and all these are contained in $[0,1]$ as a consequence of inequalities (\ref{ineq01}), (\ref{ineq02}) and (\ref{ineq03}). Consider $P(\{2,3\}) = \mu_{x}(A)+\mu_{x}(B)-2\mu_{x}(A\ {\rm and}\ B)$. From inequality (\ref{ineq03}) it follows that $\mu_{x}(A)+\mu_{x}(B)-2\mu_{x}(A\ {\rm and}\ B) \le \mu_{x}(A)+\mu_{x}(B)-\mu_{x}(A\ {\rm and}\ B) \le 1$. Further, we have, following inequalities (\ref{ineq01}) and (\ref{ineq02}), $\mu_{x}(A\ {\rm and}\ B) \le \mu_{x}(A)$ and $\mu_{x}(A\ {\rm and}\ B) \le \mu_{x}(B)$ and hence $2\mu_{x}(A\ {\rm and}\ B) \le \mu_{x}(A) + \mu_{x}(B)$. From this it follows that $0 \le \mu_{x}(A)+\mu_{x}(B)-2\mu_{x}(A\ {\rm and}\ B)$. Hence we have proved that $P(\{2,3\}) = \mu_{x}(A)+\mu_{x}(B)-2\mu_{x}(A\ {\rm and}\ B) \in [0,1]$. We have $P(\{1,4\}) = 1-\mu_{x}(A)-\mu_{x}(B)+2\mu_{x}(A\ {\rm and}\ B) = 1 - P(\{2,3\})$ and hence $P(\{1,4\}) \in [0,1]$. We have $P(\{1,2,4\})=1-\mu_{x}(B)+\mu_{x}(A\ {\rm and}\ B) = 1-P(\{3\}) \in [0,1]$ and
$P(\{1,3,4\})=1-\mu_{x}(A)+\mu_{x}(A\ {\rm and}\ B) = 1-P(\{2\}) \in [0,1]$. The last subset to control is $\Omega$ itself. We have $P(\Omega)=P(\{1\}) + P(\{2\}) + P(\{3\}) + P(\{4\}) = 1$. We have verified all subsets $S \subseteq \Omega$, and hence proved that $P$ is a probability measure. Since $P(\{1\}) = \mu_{x}(A\ {\rm and}\ B)$, $P(\{1,2\})=\mu_{x}(A)$ and $P(\{1,3\})=\mu_{x}(B)$, we have modeled the data $\mu_{x}(A)$, $\mu_{x}(B)$ and $\mu_{x}(A\ {\rm and}\ B)$ by means of a Kolmogorovian probability space, and hence they are classical conjunction data.
\qed

\medskip
\noindent
Inequalities (\ref{ineq01}) and (\ref{ineq02}) hold if and only if the quantity $\Delta_{AB}(x)=\mu_{x}(A\ {\rm and}\ B) - {\rm min}(\mu_{x}(A), \mu_{x}(B))\le 0$. The quantity $\Delta_{AB}(x)$ is called the `conjunction minimum rule deviation', since it expresses compatibility with the `minimum rule for the conjunction' in fuzzy set theory. A situation where $\Delta_{AB}(x)>0$ was called `overextension' by Hampton (1988a). The quantity $k_{AB}(x)=1-\mu_{x}(A)-\mu_{x}(B)+\mu_{x}(A \ {\rm and} \ B)$ is called the `Kolmogorovian conjunction factor'. Its violation is due to a non-classicality that is different from the one entailing the violation $\Delta_{AB}(x)$ (Aerts, 2009a). Finally, let us introduce the quantity ${\rm Doub}_{AB}(x)={\rm max}(\mu_{x}(A), \mu_{x}(B))-\mu_{x}(A\ {\rm and}\ B)$. A situation where  ${\rm Doub}_{AB}(x)>0$ was called `double overextension' by Hampton (1988a). The values of the parameters $\Delta_{AB}(x)$, $k_{AB}(x)$ and ${\rm Doub}_{AB}(x)$ for our experiment are reported in Tables 1a, 2a, 3a and 4a.

Let us then come to the representability of a set of experimental data on a concept and its negation in a classical Kolmogorovian probability model.

\medskip
\noindent
\textbf{Definition 4.}
{\it We say that the membership weights $\mu_{x}(B)$ and $\mu_{x}({\rm not} \ B)$ of the exemplar $x$ with respect to the concept $B$ and its negation `${\rm not} \ B$', respectively, can be represented in a classical Kolmogorovian probability model if there exists a Kolmogorovian probability space $(\Omega,\sigma(\Omega),P)$ and an event $E_B \in \sigma(\Omega)$ of the events algebra $\sigma(\Omega)$ such that
\begin{equation} \label{definitionnegation}
P(E_B) = \mu_{x}(B) \quad P(\Omega \setminus E_{B}) = \mu_{x}({\rm not} \ B)
\end{equation}}
Analogously to the conjunction case, we can prove a useful theorem on the representability of the membership weights with respect to a positive concept $A$, a negated concept `${\rm not} \ B$' and their conjunction `$A \ {\rm and} \ {\rm not} \ B$'.

\medskip
\noindent
\textbf{Theorem 2.} {\it The membership weights $\mu_{x}(A)$, $\mu_{x}(B)$, $\mu_{x}({\rm not} \ B)$ and $\mu_{x}(A\ {\rm and} \ {\rm not} \ B)$ of the exemplar $x$ with respect to the pair of concepts $A$, $B$, the negation `${\rm not} \ B$' and the conjunction `$A \ \rm{and} \ {\rm not} \ B$', respectively,  can be represented in a classical Kolmogorovian probability model if and only if they satisfy the following inequalities:}
\begin{eqnarray}
&0 \le \mu_{x}(A\ {\rm and} \ {\rm not} \ B) \le \mu_{x}(A) \le 1 \label{ineq04} \\
&0 \le \mu_{x}(A\ {\rm and}\ {\rm not} \ B) \le \mu_{x}({\rm not} \ B) \le 1 \label{ineq05} \\
& \mu_{x}(A) + \mu_{x}({\rm not} \ B) - \mu_{x}(A\ {\rm and}\ {\rm not} \ B) \le 1 \label{ineq06} \\
&1-\mu_{x}(B)-\mu_{x}({\rm not} \ B)=0 \label{ineq07}
\end{eqnarray}
\textbf{Proof.}
If $\mu_{x}(A), \mu_{x}({\rm not} \ B)$ and $\mu_{x}(A\ {\rm and}\ {\rm not} \ B)$ can be represented in a classical probability model, then there exists a Kolmogorovian probability space $(\Omega,\sigma(\Omega),P)$ and events $E_A, E_B \in \sigma(\Omega)$ such that $P(E_A) = \mu_{x}(A)$, $P(\Omega \setminus E_B) = \mu_{x}({\rm not} \ B)$ and $P(E_A \cap (\Omega \setminus E_B)) = \mu_{x}(A\ {\rm and}\ {\rm not} \ B)$. From the general properties of a Kolmogorovian probability space it follows that we have $0 \le P(E_A \cap (\Omega \setminus E_B)) \le P(E_A) \le 1$ and $0 \le P(E_A \cap (\Omega \setminus E_B)) \le P(\Omega \setminus E_B) \le 1$, which proves that inequalities (\ref{ineq04}) and (\ref{ineq05}) are satisfied. From the same general properties of a Kolmogorovian probability space it also follows that we have $P(E_A \cup (\Omega \setminus E_B)) = P(E_A) + P(\Omega \setminus E_B) - P(E_A \cap (\Omega \setminus E_B))$, and since $P(E_A \cup (\Omega \setminus E_B)) \le 1$ we also have $P(E_A) + P(\Omega \setminus E_B) - P(E_A \cap (\Omega \setminus E_B)) \le 1$. This proves that inequality (\ref{ineq06}) is satisfied. Finally, we have $P(E_{B})+P(\Omega \setminus E_{B})=P(\Omega)=1$, in a Kolmogorovian probability space, which proves that inequality (\ref{ineq07}) is satisfied. We have now proved that for the classical conjunction data $\mu_{x}(A), \mu_{x}(B)$, $\mu_{x}({\rm not} \ B)$ and $\mu_{x}(A\ {\rm and}\ {\rm not} \ B)$ the three inequalities are satisfied.

Now suppose that we have an exemplar $x$ whose membership weights $\mu_{x}(A), \mu_{x}({\rm not} \ B), \mu_{x}(A\ {\rm and}\ {\rm not} \ B)$ with respect to the concepts $A$ and `${\rm not} \ B$' and their conjunction `$A$ and ${\rm not} \ B$' are such that inequalities (\ref{ineq04}), (\ref{ineq05}), (\ref{ineq06}) and (\ref{ineq07}) are satisfied. We prove that, as a consequence, $\mu_{x}(A), \mu_{x}({\rm not} \ B)$ and $\mu_{x}(A\ {\rm and}\ {\rm not} \ B)$ can be represented in a Kolmogorovian probability model. To this end we explicitly construct a Kolmogorovian probability space that models these data.
Consider the set $\Omega=\{1, 2, 3, 4\}$ and $\sigma(\Omega) = {\cal P}(\Omega)$, the set of all subsets of $\Omega$. We define
\begin{eqnarray}
&P(\{1\}) = \mu_{x}(A\ {\rm and}\ {\rm not} \ B) \\ \label{eqtheorem202}
&P(\{2\}) = \mu_{x}(A) - \mu_{x}(A\ {\rm and}\ {\rm not} \ B) \\ \label{eqtheorem203}
&P(\{3\}) = \mu_{x}({\rm not} \ B) - \mu_{x}(A\ {\rm and}\ {\rm not} \ B) \\ \label{eqtheorem204}
&P(\{4\}) = 1-\mu_{x}(A)-\mu_{x}({\rm not} \ B)+\mu_{x}(A\ {\rm and}\ {\rm not} \ B)
\end{eqnarray}
and further for an arbitrary subset $S \subseteq \{1,2,3,4\}$ we define
\begin{equation} \label{defarbitrarysubset02}
P(S) = \sum_{a\in S}P(\{a\})
\end{equation}
Let us prove that $P: \sigma(\Omega) \rightarrow [0,1]$ is a probability measure. To this end we need to prove that $P(S) \in [0,1]$ for an arbitrary subset $S \subseteq \Omega$, and that the `sum formula' for a probability measure is satisfied to comply with (\ref{KolmogorovianMeasure}). The sum formula for a probability measure is satisfied because of definition (\ref{defarbitrarysubset02}). What remains to be proved is that $P(S) \in [0,1]$ for an arbitrary subset $S \subseteq \Omega$. $P(\{1\}), P(\{2\}), P(\{3\})$ and $P(\{4\})$ are contained in $[0,1]$ as a direct consequence of inequalities (\ref{ineq04}), (\ref{ineq05}) and (\ref{ineq06}). Further, we have $P(\{1,2\}) = \mu_{x}(A), P(\{1,3\})=\mu_{x}({\rm not} \ B), P(\{3,4\}) = 1-\mu_{x}(A), P(\{2,4\}) = 1-\mu_{x}({\rm not} \ B)=\mu_{x}(B)$, because of Equation (\ref{ineq07}), $P(\{2,3,4\}) = 1-\mu_{x}(A \ {\rm and} \ {\rm not} \ B)$ and $P(\{1,2,3\}) = \mu_{x}(A)+\mu_{x}({\rm not} \ B)-\mu_{x}(A\ {\rm and}\ {\rm not} \ B)$, and all these are contained in $[0,1]$ as a consequence of inequalities (\ref{ineq04}), (\ref{ineq05}) and (\ref{ineq06}). Consider $P(\{2,3\}) = \mu_{x}(A)+\mu_{x}({\rm} \ B)-2\mu_{x}(A\ {\rm and}\ {\rm not} \ B)$. From inequality (\ref{ineq06}) it follows that $\mu_{x}(A)+\mu_{x}({\rm not} \ B)-2\mu_{x}(A\ {\rm and}\ {\rm not} \ B) \le \mu_{x}(A)+\mu_{x}({\rm not} \ B)-\mu_{x}(A\ {\rm and}\ {\rm not} \ B) \le 1$. Further, we have, following inequalities (\ref{ineq04}) and (\ref{ineq05}), $\mu_{x}(A\ {\rm and}\ {\rm not} \ B) \le \mu_{x}(A)$ and $\mu_{x}(A\ {\rm and}\ {\rm not} \ B) \le \mu_{x}({\rm not} \ B)$ and hence $2\mu_{x}(A\ {\rm and}\ {\rm not} \ B) \le \mu_{x}(A) + \mu_{x}({\rm not} \ B)$. From this it follows that $0 \le \mu_{x}(A)+\mu_{x}({\rm not} \ B)-2\mu_{x}(A\ {\rm and}\ {\rm not} \ B)$. Hence we have proved that $P(\{2,3\}) = \mu_{x}(A)+\mu_{x}({\rm not} \ B)-2\mu_{x}(A\ {\rm and}\ {\rm not} \ B) \in [0,1]$. We have $P(\{1,4\}) = 1-\mu_{x}(A)-\mu_{x}({\rm not} \ B)+2\mu_{x}(A\ {\rm and}\ {\rm not} \ B) = 1 - P(\{2,3\})$ and hence $P(\{1,4\}) \in [0,1]$. We have $P(\{1,2,4\})=1-\mu_{x}({\rm not} \ B)+\mu_{x}(A\ {\rm and}\ {\rm not} \  B) = 1-P(\{3\}) \in [0,1]$ and
$P(\{1,3,4\})=1-\mu_{x}(A)+\mu_{x}(A\ {\rm and}\ {\rm not} \ B) = 1-P(\{2\}) \in [0,1]$. The last subset to control is $\Omega$ itself. We have $P(\Omega)=P(\{1\}) + P(\{2\}) + P(\{3\}) + P(\{4\}) = 1$. We have verified all subsets $S \subseteq \Omega$, and hence proved that $P$ is a probability measure. Since $P(\{1\}) = \mu_{x}(A\ {\rm and}\ {\rm not} \  B)$, $P(\{1,2\})=\mu_{x}(A)$, $P(\{1,3\})=\mu_{x}({\rm not} \ B)$, $P(\{2,4\})=\mu_{x}(B)$ and $P(\Omega \setminus \{ 2,4\})=\mu_{x}({\rm not} \ B)$, we have modeled the data $\mu_{x}(A)$, $\mu_{x}(B)$ and $\mu_{x}(A\ {\rm and}\ B)$ by means of a Kolmogorovian probability space, and hence they are classical conjunction data.
\qed

\medskip
\noindent
Inequalities (\ref{ineq04}) and (\ref{ineq05}) hold if and only if the conjunction minimum rule deviation $\Delta_{AB'}(x)=\mu_{x}(A\ {\rm and}\ {\rm not} \ B) - {\rm min}(\mu_{x}(A), \mu_{x}({\rm not} \ B))\le 0$. A situation where $\Delta_{AB'}(x)>0$ entails that `overextension' is present. The Kolmogorovian conjunction factor $k_{AB'}(x)=1-\mu_{x}(A)-\mu_{x}({\rm not} \ B)+\mu_{x}(A \ {\rm and} \ {\rm not} \  B) \le 0$ and the quantity $l_{BB'}(x)=1-\mu_{x}(B)-\mu_{x}({\rm not} \ B)=0$ complete the classicality of the conjunction `$A \ {\rm and} \ {\rm not} \ B$'. Then, let us introduce the quantity ${\rm Doub}_{AB'}(x)={\rm max}(\mu_{x}(A), \mu_{x}({\rm not} \ B))-\mu_{x}(A\ {\rm and}\  {\rm not} \ B)$. A situation where  ${\rm Doub}_{AB'}(x)>0$ is a situation of `double overextension'. The quantity $l_{BB'}(x)=1-\mu_{x}(B)-\mu_{x}({\rm not} \ B)$ in Equation (\ref{ineq07}) is a new parameter that must be introduced to represent the negation of a concept in terms of a classical set-theoretic complementation, namely,. A situation where $l_{BB'}(x)\ne 0$ produces a type of deviation from classicality and it is due to conceptual negation. The values of the parameters $\Delta_{AB}(x)$, $k_{AB}(x)$, ${\rm Doub}_{AB}(x)$ and $l_{BB'}(x)$ for our experiment are reported in Tables 1b, 2b, 3b and 4b.

Let us now come back to our experiments. Theorems 1 and 2 are manifestly violated by several exemplars with respect to both conjunctions `$A \ {\rm and} \ B$' and `$A \ {\rm and} \ {\rm not} \ B$'. It is however interesting to observe that the conditions $k_{AB}(x)>0$ and $k_{AB'}(x)>0$ are never violated, hence the deviations from a classical probability model in our experimental data are all due to overextension in the conjunctions `$A \ {\rm and} \ B$' and `$A \ {\rm and} \ {\rm not} \ B$' and to a violation of $l_{BB'}=0$ in the negation ${\rm not} \ B$. For example, the item {\it Prize Bull} has $\Delta_{AB}(x)=0.29375>0$ with respect to {\it Pets And Farmyard Animals}, hence it is strongly overextended with respect to {\it Pets And Farmyard Animals}, and it is even double overextended with ${\rm Doub}_{AB'}(x)=-0.0125<0$ with respect to {\it Pets and Not Farmyard Animals}. The already mentioned {\it Broccoli} and {\it Parsley} are such that their $\Delta_{AB}(x)$s are equal to $0.43125$ and $0.49375$, respectively. The exemplar {\it Chili Pepper} has $\Delta_{AB'}(x)=0.3375$, while the exemplar {\it Broccoli} has $\Delta_{AB'}(x)=0.31875$, both with respect to {\it Fruits And Not Vegetables}, hence they are both highly overextended.

It is finally interesting to observe that evident deviations from classicality are also due to conceptual negation. Let us consider some cases. The exemplar {\it Rug} has $l_{BB'}(x)=-0.18125$ with respect to the concept {\it Furniture} and its negation {\it Not Furniture}, while {\it Wall Mirror} has $l_{BB'}(x)=-0.20625$ with respect to the same concept and negation. Other relevant examples are {\it Sugar} and {\it Chives} with $l_{BB'}(x)=-0.1125$ and $l_{BB'}(x)=-0.14375$, respectively, with respect to {\it Herbs} and {\it Not Herbs}, and {\it Collie Dog} with $l_{BB'}(x)=-0.1188$ with respect to {\it Farmyard Animals} and {\it Not Farmyard Animals}.

The results obtained in this section point to a systematic deviation of our experimental data from the rules of classical (fuzzy set) logic and probability theory. It is then worth to investigate whether the `non-classicalities' identified here are of a quantum-type, and hence they can be described within the mathematical formalism of quantum theory. To this end we need to preliminary summarize the essentials of the quantum mathematics that is needed to employ this quantum formalism for modeling purposes.

\section{Fundamentals of a quantum-theoretic modeling\label{quantum}}
We illustrate in this section how the mathematical formalism of quantum theory can be applied to model situations outside the microscopic quantum world, more specifically, in the representation of concepts and their combinations. We avoid in our presentation superfuous technicalities, but aim to be synthetic and rigorous at the same time.

When the quantum mechanical formalism is applied for modeling purposes, each considered entity  -- in our case a concept -- is associated with a complex Hilbert space ${\cal H}$, that is, a vector space over the field ${\mathbb C}$ of complex numbers, equipped with an inner product $\langle \cdot |  \cdot \rangle$ that maps two vectors $\langle A|$ and $|B\rangle$ onto a complex number $\langle A|B\rangle$. We denote vectors by using the bra-ket notation introduced by Paul Adrien Dirac, one of the pioneers of quantum theory (Dirac, 1958). Vectors can be `kets', denoted by $\left| A \right\rangle $, $\left| B \right\rangle$, or `bras', denoted by $\left\langle A \right|$, $\left\langle B \right|$. The inner product between the ket vectors $|A\rangle$ and $|B\rangle$, or the bra-vectors $\langle A|$ and $\langle B|$, is realized by juxtaposing the bra vector $\langle A|$ and the ket vector $|B\rangle$, and $\langle A|B\rangle$ is also called a `bra-ket', and it satisfies the following properties:

(i) $\langle A |  A \rangle \ge 0$;

(ii) $\langle A |  B \rangle=\langle B |  A \rangle^{*}$, where $\langle B |  A \rangle^{*}$ is the complex conjugate of $\langle A |  B \rangle$;

(iii) $\langle A |(z|B\rangle+t|C\rangle)=z\langle A |  B \rangle+t \langle A |  C \rangle $, for $z, t \in {\mathbb C}$,
where the sum vector $z|B\rangle+t|C\rangle$ is called a `superposition' of vectors $|B\rangle$ and $|C\rangle$ in the quantum jargon.

From (ii) and (iii) follows that inner product $\langle \cdot |  \cdot \rangle$ is linear in the ket and anti-linear in the bra, i.e. $(z\langle A|+t\langle B|)|C\rangle=z^{*}\langle A | C\rangle+t^{*}\langle B|C \rangle$.

We recall that the `absolute value' of a complex number is defined as the square root of the product of this complex number times its complex conjugate, that is, $|z|=\sqrt{z^{*}z}$. Moreover, a complex number $z$ can either be decomposed into its cartesian form $z=x+iy$, or into its goniometric form $z=|z|e^{i\theta}=|z|(\cos\theta+i\sin\theta)$.  As a consequence, we have $|\langle A| B\rangle|=\sqrt{\langle A|B\rangle\langle B|A\rangle}$. We define the `length' of a ket (bra) vector $|A\rangle$ ($\langle A|$) as $|| |A\rangle ||=||\langle A |||=\sqrt{\langle A |A\rangle}$. A vector of unitary length is called a `unit vector'. We say that the ket vectors $|A\rangle$ and $|B\rangle$ are `orthogonal' and write $|A\rangle \perp |B\rangle$ if $\langle A|B\rangle=0$.

We have now introduced the necessary mathematics to state the first modeling rule of quantum theory, as follows.

\medskip
\noindent{\it First quantum modeling rule:} A state $A$ of an entity -- in our case a concept -- modeled by quantum theory is represented by a ket vector $|A\rangle$ with length 1, that is $\langle A|A\rangle=1$.

\medskip
\noindent
An orthogonal projection $M$ is a linear operator on the Hilbert space, that is, a mapping $M: {\cal H} \rightarrow {\cal H}, |A\rangle \mapsto M|A\rangle$ which is Hermitian and idempotent. The latter means that, for every $|A\rangle, |B\rangle \in {\cal H}$ and $z, t \in {\mathbb C}$, we have:

(i) $M(z|A\rangle+t|B\rangle)=zM|A\rangle+tM|B\rangle$ (linearity);

(ii) $\langle A|M|B\rangle=\langle B|M|A\rangle$ (hermiticity);

(iii) $M \cdot M=M$ (idempotency).

The identity operator $\mathbbmss{1}$ maps each vector onto itself and is a trivial orthogonal projection. We say that two orthogonal projections $M_k$ and $M_l$ are orthogonal operators if each vector contained in $M_k({\cal H})$ is orthogonal to each vector contained in $M_l({\cal H})$, and we write $M_k \perp M_l$, in this case. The orthogonality of the projection operators $M_{k}$ and $M_{l}$ can also be expressed by $M_{k}M_{l}=0$, where $0$ is the null operator. A set of orthogonal projection operators $\{M_k\ \vert k=1,\ldots,n\}$ is called a `spectral family' if all projectors are mutually orthogonal, that is, $M_k \perp M_l$ for $k \not= l$, and their sum is the identity, that is, $\sum_{k=1}^nM_k=\mathbbmss{1}$.

The above definitions give us the necessary mathematics to state the second modeling rule of quantum theory, as follows.

\medskip
\noindent
{\it Second quantum modeling rule:} A measurable quantity $Q$ of an entity -- in our case a concept -- modeled by quantum theory, and having a set of possible real values $\{q_1, \ldots, q_n\}$ is represented by a spectral family $\{M_k\ \vert k=1, \ldots, n\}$ in the following way. If the entity -- in our case a concept -- is in a state represented by the vector $|A\rangle$, then the probability of obtaining the value $q_k$ in a measurement of the measurable quantity $Q$ is $\langle A|M_k|A\rangle=||M_k |A\rangle||^{2}$. This formula is called the `Born rule' in the quantum jargon. Moreover, if the value $q_k$ is actually obtained in the measurement, then the initial state is changed into a state represented by the vector
\begin{equation}
|A_k\rangle=\frac{M_k|A\rangle}{||M_k|A\rangle||}
\end{equation}
This change of state is called `collapse' in the quantum jargon.

\medskip
\noindent
The tensor product ${\cal H}_{A} \otimes {\cal H}_{B}$ of two Hilbert spaces ${\cal H}_{A}$ and ${\cal H}_{B}$ is the Hilbert space generated by the set $\{|A_i\rangle \otimes |B_j\rangle\}$, where $|A_i\rangle$ and $|B_j\rangle$ are vectors of ${\cal H}_{A}$ and ${\cal H}_{B}$, respectively, which means that a general vector of this tensor product is of the form $\sum_{ij}|A_i\rangle \otimes |B_j\rangle$. This gives us the necessary mathematics to introduce the third modeling rule.

\medskip
\noindent
{\it Third quantum modeling rule:} A state $C$ of a compound entity -- in our case a combined concept -- is represented by a unit vector $|C\rangle$ of the tensor product ${\cal H}_{A} \otimes {\cal H}_{B}$ of the two Hilbert spaces ${\cal H}_{A}$ and ${\cal H}_{B}$ containing the vectors that represent the states of the component entities -- concepts.

\medskip
\noindent
The above means that we have $|C\rangle=\sum_{ij}c_{ij}|A_i\rangle \otimes |B_j\rangle$, where $|A_i\rangle$ and $|B_j\rangle$ are unit vectors of ${\cal H}_{A}$ and ${\cal H}_{B}$, respectively, and $\sum_{i,j}|c_{ij}|^{2}=1$. We say that the state $C$ represented by $|C\rangle$ is a product state if it is of the form $|A\rangle \otimes |B\rangle$ for some $|A\rangle \in {\cal H}_{A}$ and $|B\rangle \in {\cal H}_{B}$. Otherwise, $C$ is called an `entangled state'.

\medskip
\noindent
The Fock space is a specific type of Hilbert space, originally introduced in quantum field theory. For most states of a quantum field the number of identical quantum entities is not conserved but is a variable quantity. The Fock space copes with this situation in allowing its vectors to be superpositions of vectors pertaining to different sectors for fixed numbers of identical quantum entities. More explicitly, the $k$-th sector of a Fock space describes a fixed number of $k$ identical quantum entities, and it is of the form ${\cal H}\otimes \ldots \otimes{\cal H}$ of the tensor product of $k$ identical Hilbert spaces ${\cal H}$. The Fock space $F$ itself is the direct sum of all these sectors, hence
\begin{equation} \label{fockspace}
{\cal F}=\oplus_{k=1}^j\otimes_{l=1}^k{\cal H}
\end{equation}
For our modeling we have only used Fock space for the `two' and `one quantum entity' case, hence ${\cal F}={\cal H}\oplus({\cal H}\otimes{\cal H})$. This is due to considering only combinations of two concepts. The sector ${\cal H}$ is called the `sector 1', while the sector ${\cal H}\otimes{\cal H}$ is called the `sector 2'. A unit vector $|F\rangle \in {\cal F}$ is then written as $|F\rangle = ne^{i\gamma}|C\rangle+me^{i\delta}(|A\rangle\otimes|B\rangle)$, where $|A\rangle, |B\rangle$ and $|C\rangle$ are unit vectors of ${\cal H}$, and such that $n^2+m^2=1$. For combinations of $j$ concepts, the general form of Fock space expressed in Equation (\ref{fockspace}) will have to be used.

This quantum-theoretic modeling can be generalized by allowing states to be represented by the so called `density operators' and measurements to be represented by the so called `positive operator valued measures'. However, our representation above is sufficient for attaining the results in this paper and we will use it in the following sections.

\section{A quantum model for the combination of two concepts\label{brussels}}
In this section, we put forward the quantum-theoretic framework that has been employed to model Hampton's (Hampton, 1988a,b) and Alxatib \& Pelletier's (Alxatib \& Pelletier, 2011), applying it to our experiment reported in Section \ref{experiment}. We show that this framework, once specified for the given conceptual combinations, i.e. `$A \ \textrm{and} \ B$' and `$A \ \textrm{and} \ \textrm{not} B$', enables a complete and successful modeling of those experimental data collected in Section \ref{experiment}.

Let us start from the disjunction `$A \ \textrm{or} \ B$' of two concepts $A$ and $B$. When the membership of the exemplar (item) $x$ with respect to $A$ is measured, we represent $A$ by the unit vector $|A_{d}(x)\rangle$ of a Hilbert space $\cal H$, and describe the decision measurement of a subject estimating whether $x$ is a member of $A$ by means of a dichotomic observable represented by the orthogonal projection operator $M$. The probability $\mu_{x}(A)$ that $x$ is chosen as a member of $A$, i.e. its membership weight, is given by the scalar product $\mu_{x}(A)=\langle A_{d}(x) | M|A_{d}(x)\rangle$. Let $A$ and $B$ be two concepts, represented by the unit vectors $|A_{d}(x)\rangle$ and $|B_{d}(x)\rangle$, respectively. To represent the concept `$A \ \textrm{or} \ B$' we take the archetypical situation of the quantum double slit experiment, where $|A\rangle$ and $|B\rangle$ represent the states of a quantum particle in which only one slit is open, ${1 \over \sqrt{2}}(|A\rangle+|B\rangle)$ represents the state of the quantum particle in which both slits are open, and $\mu_{x}(A\ {\rm or}\ B)$ is the probability that the quantum particle is detected in a given region 
of a screen behind the slits. Thus, the concept `$A \ \textrm{or} \ B$' is represented by the unit vector ${1 \over \sqrt{2}}(|A_{d}(x)\rangle+|B_{d}(x)\rangle)$, and  $|A_{d}(x)\rangle$ and $|B_{d}(x)\rangle$ are chosen to be orthogonal, that is, $\langle A_{d}(x) | B_{d}(x) \rangle=0$. The membership weights $\mu_{x}(A), \mu_{x}(B)$ and $\mu_{x}(A\ {\rm or} \ B)$ of an exemplar $x$ for the concepts $A$, $B$ and `$A \ {\rm or} \ B$' are given by
\begin{eqnarray}
\mu_{x}(A)&=&\langle A_{d}(x) | M|A_{d}(x)\rangle \\
\mu_{x}(B)&=&\langle B_{d}(x) | M|B_{d}(x)\rangle \\
\mu_{x}(A \ {\rm or} \ B)&=&{1 \over 2}(\mu_{x}(A)+\mu_{x}(B))+\Re\langle A_{d}(x)|M|B_{d}(x)\rangle
\end{eqnarray}
repsectively, where $\Re\langle A_{d}(x)|M|B_{d}(x)\rangle$ is the real part of the complex number $\langle A_{d}(x)|M|B_{d}(x)\rangle$. The complex term $\Re\langle A_{d}(x)|M|B_{d}(x)\rangle$ is called `interference term' in the quantum jargon, since it produces a deviation from the average ${1 \over 2}(\mu_{x}(A)+\mu_{x}(B))$ which would have been observed in the quantum double slit experiment in absence of interference. We can see that, already at this stage, two genuine quantum effects, namely, superposition and interference, occur in the mechanism of combination of the concepts $A$ and $B$.

The quantum-theoretic modeling presented above correctly describes a large part of data in Hampton (1988b), but it cannot cope with quite some cases -- in fact most of all the cases that behave more classically than the ones that are easily modeled by quantum interference. The reason is that, if one wants to reproduce Hampton's data within a quantum mathematics model which fully exploits the analogy with the quantum double slit experiment, one has to include the situation in which two identical quantum particles are considered, both particles passes through the slits, and the probability that at least one particle is detected in the spot $x$ is calculated. This probability is given by $\mu_{x}(A)+\mu_{x}(B)-\mu_{x}(A)\mu_{x}(B)$ (Aerts, 2009a). Quantum field theory in Fock space allows one to complete the model, as follows.

In quantum field theory, a quantum entity is described by a field which consists of superpositions of different configurations of many quantum particles (see Section \ref{quantum}). Thus, the quantum entity is associated with a Fock space $\cal F$ which is the direct sum $\oplus$ of different Hilbert spaces, each Hilbert space describing a defined number of quantum particles. In the simplest case, ${\cal F}={\cal H} \oplus ({\cal H}\otimes {\cal H})$, where $\cal H$ is the Hilbert space of a single quantum particle (sector 1 of $\cal F$) and ${\cal H} \otimes {\cal H}$ is the (tensor product) Hilbert space of two identical quantum particles (sector 2 of $\cal F$).

Let us come back to our modeling for concept combinations. The normalized superposition ${1 \over \sqrt{2}}(|A_{d}(x)\rangle+|B_{d}(x)\rangle)$ represents the state of the new emergent concept `$A \ \textrm{or} \ B$' in sector 1 of the Fock space $\cal F$. In sector 2 of $\cal F$, instead, the state of the concept `$A \ \textrm{or} \ B$' is represented by the unit (product) vector $|A_{d}(x)\rangle\otimes|B_{d}(x)\rangle$. To describe the decision measurement in this sector, we first suppose that the subject considers two identical copies of the exemplar $x$, pondering on the membership of the first copy of $x$ with respect to $A$ `and' the membership of the second copy of $x$ with respect to $B$. The probability of getting `yes' in both cases is, by using quantum mechanical rules, $(\langle A_{d}(x)|\langle B_{d}(x)|)|M \otimes M | (|A_{d}(x)\rangle\otimes|B_{d}(x)\rangle)$. The probability of getting at least a positive answer is instead $1-(\langle A_{d}(x)|\langle B_{d}(x)|)|(\mathbbmss{1}-M) \otimes (\mathbbmss{1}- M) | (|A_{d}(x)\rangle\otimes|B_{d}(x)\rangle)$. Hence, the membership weight of the exemplar $x$ with respect to the concept `$A \ \textrm{or} \ B$' coincides in sector 2 with the latter probability and can be written as  $1-(\langle A_{d}(x)|\langle B_{d}(x)|)|(\mathbbmss{1}-M) \otimes (\mathbbmss{1}- M) | (|A_{d}(x)\rangle\otimes|B_{d}(x)\rangle)=\mu_{x}(A)+\mu_{x}(B)-\mu_{x}(A)\mu_{x}(B)=(\langle A_{d}(x)|\langle B_{d}(x)|)| M \otimes \mathbbmss{1}+\mathbbmss{1}\otimes M - M \otimes M | (|A_{d}(x)\rangle\otimes|B_{d}(x)\rangle)$.

Coming to the Fock space ${\cal F}={\cal H} \oplus ({\cal H}\otimes {\cal H})$, the global initial state of the concepts is represented by the unit vector
\begin{equation}
|A \ \textrm{or} \ B(x)\rangle=m_{d}(x) e^{i\lambda_{d}(x)}|A_{d}(x)\rangle\otimes|B_{d}(x)\rangle+n_{d}(x)e^{i\nu_{d}(x)}{1\over \sqrt{2}}(|A_{d}(x)\rangle+|B_{d}(x)\rangle)
\end{equation}
where the real numbers $m_{d}(x),n_{d}(x)$ are such that $0\le m_{d}(x),n_{d}(x)$ and $m_{d}(x)^2+n_{d}(x)^2=1$. The decision measurement on the membership of the exemplar $x$ with respect to the concept `$A \ \textrm{or} \ B$' is represented by the orthogonal projection operator $M \oplus (M \otimes \mathbbmss{1}+\mathbbmss{1}\otimes M - M \otimes M)$, hence the membership weight of $x$ with respect to `$A \ \textrm{or} \ B$' is given by
\begin{eqnarray}
\mu_{x}(A \ \textrm{or} \ B)&=&\langle  A \ \textrm{or} \ B(x) |M \oplus (M \otimes \mathbbmss{1}+\mathbbmss{1}\otimes M - M \otimes M)| A \ \textrm{or} \ B(x) \rangle \nonumber \\ &=&m_{d}(x)^2(\mu_{x}(A)+\mu_{x}(B)-\mu_{x}(A)\mu_{x}(B))+n_{x}(x)^2({\mu_{x}(A)+\mu_{x}(B) \over 2}+\Re\langle A_{d}(x)|M|B_{d}(x)\rangle) \label{OR}
\end{eqnarray}
The simplest Fock space that allows the modeling of `$A \ \textrm{or} \ B$' is ${\mathbb C}^{3}\oplus ({\mathbb C}^{3}\otimes{\mathbb C}^{3})$. Let us denote by $|1,0,0\rangle,|0,1,0\rangle,|0,0,1\rangle$ the canonical basis of ${\mathbb C}^{3}$. Then, let us set $a_{x}(A)=1-\mu_{x}(A)$ and $b_{x}(B)=1-\mu_{x}(B)$ if $\mu_{x}(A)+\mu_{x}(B)\le 1$, $a_{x}(A)=\mu_{x}(A)$ and $b_{x}(B)=\mu_{x}(B)$ if $\mu_{x}(A)+\mu_{x}(B) > 1$. In Aerts (2009a) and Aerts, Gabora \& Sozzo (2012) it has been proved that, independently of the value of $\mu_{x}(A)+\mu_{x}(B)$, the interference term $\Re\langle A_{d}(x)|M|B_{d}(x)\rangle$ is given by
\begin{equation} \label{intD}
\Re\langle A_{d}(x)|M|B_{d}(x)\rangle=\sqrt{1-a_{x}(A)}\sqrt{1-b_{x}(B)}\cos\theta_{d}(x)
\end{equation}
where $\theta_{d}(x)$ is the `interference angle'. The unit vectors $|A_{d}(x)\rangle$ and $|B_{d}(x)\rangle$ are instead represented in the canonical basis of ${\mathbb C}^{3}$ by
\begin{eqnarray}
|A_{d}(x)\rangle&=&\Big ( \sqrt{a_{x}(A)}, 0, \sqrt{1-a_{x}(A)} \Big ) \\
|B_{d}(x)\rangle&=&e^{i \theta_{d}(x)}\Big ( \sqrt{\frac{(1-a_{x}(A))(1-b_{x}(B))}{a_x(A)}}, \sqrt{\frac{a_{x}(A)+b_{x}(B)-1}{a_{x}(A)}}, -\sqrt{1-b_{x}(B)} \Big ) \quad {\rm if} \quad a_{x}(A)\ne 0  \\
|B_{d}(x)\rangle&=&e^{i \theta_{d}(x)}(0,1,0) \quad {\rm if} \quad a_{x}(A)= 0
\end{eqnarray}
and the interference angle satisfies the condition
\begin{equation}
\theta_{d}(x)= \arccos \Big ( \frac{
{2 \over n_{d}(x)^2} \Big (\mu_{x}(A \ \textrm{or} \ B)-m_{d}(x)^2 (1-\mu_{x}(A)-\mu_{x}(B)-\mu_{x}(A)\mu_{x}(B)) \Big )-\mu_{x}(A)-\mu_{x}(B)}{\sqrt{1-a_x(A)}\sqrt{1-b_x(B)}} \Big )
\end{equation}
if $a_{x}(A), b_{x}(B)\ne 1$. The angle $\theta_{d}(x)$ is instead arbitrary if $a_{x}(A)=1$ or $b_{x}(B)=1$ (Aerts, 2009a; Aerts, Gabora \& Sozzo, 2013).

Let us now come to the representation for the conjunction `$A \ \textrm{and} \ B$'. Here, the decision measurement for the membership weight of the exemplar $x$ with respect to the concept `$A \ \textrm{and} \ B$' is represented in the Fock space ${\cal F}={\cal H} \oplus ({\cal H}\otimes {\cal H})$ by the orthogonal projection operator $M \oplus (M \otimes M)$, while the membership weight of $x$ with respect to `$A \ \textrm{and} \ B$' is given by\footnote{The membership weight $\mu_{x}(A \ \textrm{or} \ B)$ could have been calculated from the membership weight $\mu_{x}(A \ \textrm{and} \ B)$ by observing that the probability that a subject decides for the membership of the exemplar $x$ with respect to the concept `$A \ \textrm{or} \ B$' is 1 minus the probability of decision against membership of $x$ with respect to the concept `$A \ \textrm{and} \ B$'.}
\begin{eqnarray} 
\mu_{x}(A \ \textrm{and} \ B)=\langle A \ \textrm{and} \ B(x) |M \oplus (M \otimes M)| A \ \textrm{and} \ B(x) \rangle \nonumber \\ =m_{c}(x)^2\mu_{x}(A)\mu_{x}(B)+n_{c}(x)^2({\mu_{x}(A)+\mu_{x}(B) \over 2}+\Re\langle A_{c}(x)|M|B_{c}(x)\rangle)\label{AND}
\end{eqnarray}
where $m_{c}(x),n_{c}(x)$ are such that $0\le m_{c}(x),n_{c}(x)$ and $m_{c}(x)^2+n_{c}(x)^2=1$. The unit vector $|A \ \textrm{and} \ B(x)\rangle$ is given by
\begin{equation} \label{vectorAND}
|A \ \textrm{and} \ B(x)\rangle=m_{c}(x) e^{i\lambda_{c}(x)}|A_{c}(x)\rangle\otimes|B_{c}(x)\rangle+n_{c}(x)e^{i\nu_{c}(x)}{1\over \sqrt{2}}(|A_{c}(x)\rangle+|B_{c}(x)\rangle)
\end{equation}
Also in this case, it has been proved that the interference term $\Re\langle A_{c}(x)|M|B_{c}(x)\rangle$ is given by
\begin{equation} \label{intC}
\Re\langle A_{c}(x)|M|B_{c}(x)\rangle=\sqrt{1-a_{x}(A)}\sqrt{1-a_{x}(B)}\cos\theta_{c}(x)
\end{equation}
in the Fock space ${\mathbb C}^{3}\oplus ({\mathbb C}^{3}\otimes{\mathbb C}^{3})$, $\theta_{c}(x)$ being the interference angle. The concepts $A_{c}(x)$ and $B_{c}(x)$ are respectively represented in the canonical basis $|1,0,0\rangle,|0,1,0\rangle,|0,0,1\rangle$ of ${\mathbb C}^{3}$ by the unit vectors
\begin{eqnarray}
|A_{c}(x)\rangle&=&\Big ( \sqrt{a_{x}(A)}, 0, \sqrt{1-a_{x}(A)} \Big ) \label{vectorA} \\
|B_{c}(x)\rangle&=&e^{i \theta_{c}(x)}\Big ( \sqrt{\frac{(1-a_{x}(A))(1-b_{x}(B))}{a_x(A)}}, \sqrt{\frac{a_{x}(A)+b_{x}(B)-1}{a_{x}(A)}}, -\sqrt{1-b_{x}(B)} \Big ) \quad {\rm if} \quad a_{x}(A)\ne 0  \label{vectorB1}\\
|B_{c}(x)\rangle&=&e^{i \theta_{c}(x)}(0,1,0) \quad {\rm if} \quad a_{x}(A)= 0 \label{vectorB2}
\end{eqnarray}
The interference angle satisfies the condition
\begin{equation}
\theta_{c}(x)= \arccos \Big ( \frac{
{2 \over n_{c}(x)^2} \Big (\mu_{x}(A \ \textrm{and} \ B)-m_{c}(x)^2\mu_{x}(A)\mu_{x}(B) \Big )-\mu_{x}(A)-\mu_{x}(B)}{\sqrt{1-a_x(A)}\sqrt{1-b_x(B)}} \Big )
\end{equation}
if $a_{x}(A),b_{x}(B)\ne 1$. The angle $\theta_{c}(x)$ is instead arbitrary if $a_{x}(A)=1$ or $b_{x}(B)=1$ (Aerts, 2009a; Aerts, Gabora \& Sozzo, 2013).

Let us finally particularize Equations (\ref{AND}), (\ref{vectorAND}) and (\ref{intC}) to the conjunctions `$A \ \textrm{and} \ B$' and `$A \ \textrm{and} \ \textrm{not} \ B$' in Section \ref{experiment}. We have
\begin{eqnarray}
|A \ \textrm{and} \ B(x)\rangle&=&m_{AB}(x) e^{i\lambda_{AB}(x)}|A(x)\rangle\otimes|B(x)\rangle+n_{AB}(x)e^{i\nu_{AB}(x)}{1\over \sqrt{2}}(|A(x)\rangle+|B(x)\rangle) \label{vectorAB} \\
|A \ \textrm{and} \ \textrm{not} \ B(x)\rangle&=&m_{AB'}(x) e^{i\lambda_{AB'}(x)}|A(x)\rangle\otimes|\textrm{not} \ B(x)\rangle+n_{AB'}(x)e^{i\nu_{AB'}(x)}{1\over \sqrt{2}}(|A(x)\rangle+|{\rm not} \ B(x)\rangle) \label{vectorAB'}
\end{eqnarray}
and
\begin{eqnarray}
\mu_{x}(A \ \textrm{and} \ B)=m_{AB}(x)^2\mu_{x}(A)\mu_{x}(B)+n_{AB}(x)^2({\mu_{x}(A)+\mu_{x}(B) \over 2}+ \nonumber \\ \sqrt{1-a_{x}(A)}\sqrt{1-b_{x}(B)}\cos\theta_{AB}(x)) \label{AB} \\
\mu_{x}(A \ \textrm{and} \ {\rm not} \ B)=m_{AB'}(x)^2\mu_{x}(A)\mu_{x}({\rm not} \ B)+n_{AB'}(x)^2({\mu_{x}(A)+\mu_{x}({\rm not} \ B) \over 2}+ \nonumber \\  \sqrt{1-a_{x}(A)}\sqrt{1-b_{x}({\rm not} \ B)}\cos\theta_{AB'}(x))  \label{AB'}
\end{eqnarray}
in the Fock space ${\mathbb C}^{3}\oplus ({\mathbb C}^{3}\otimes{\mathbb C}^{3})$.

We have thus completed our quantum mathematics representation of the concepts $A$, $B$, the negation  `${\rm not} \ B$' and the conjunctions `$A \ {\rm and} \ B$' and `$A \ {\rm and} \ {\rm not} \ B$' in Fock space. In the next section we will see how this representation works for the experimental data in Section \ref{experiment}.

\section{Representing the empirical data in Fock space\label{model}}
Equations (\ref{OR}) and (\ref{AND}) in Section \ref{brussels} contain the quantum probabilistic expressions allowing the modeling of a major part of Hampton's data (1988a,b). Moreover, we have showed in Sozzo (2014) that Equation (\ref{AND}) can also model (Alxatib \& Pelletier, 2012)'s data on borderline vagueness. We show in this section that almost all the data collected in our experiments on `$A \ {\rm and} \ B$' and `$A \ {\rm and} \ {\rm not} \ B$' can be modeled in the same Fock space framework.

Let us start from the conjunction `$A \ {\rm and} \ B$'. Tables 5a, 6a, 7a and 8a report, for each exemplar $x$, the values of the interference angle $\theta_{AB}(x)$ and the weights $m_{AB}(x)^2$ and $n_{AB}(x)^2$ which satisfy Equation (\ref{AB}), together with the representation of the unit vectors $|A_{AB}(x)\rangle$ and $|B_{AB}(x)\rangle$ in ${\mathbb C}^{3}$ satisfying Equations (\ref{vectorA}), (\ref{vectorB1}) and (\ref{vectorB2}). Let us consider the exemplar {\it Olive} which was double overextended  in Section \ref{experiment}, since it scored a membership weight $\mu_{x}(A)=0.53125$ with respect to {\it Fruits}, $\mu_{x}(B)=0.63125$ with respect to {\it Vegetables}, and $\mu_{x}(A \ {\rm and} \ B)=0.65$ with respect to {\it Fruits And Vegetables}. As we can see from Table 8a, {\it Olive} can be modeled in the Fock space ${\mathbb C}^{3}\oplus ({\mathbb C}^{3}\otimes{\mathbb C}^{3})$ with an interference angle	$\theta_{AB}(x)=60.48^{\circ}$, a weight $m_{AB}(x)^2=0.3$ in sector 2, and	a weight $n_{AB}(x)^2=0.7$ in sector 1. The concepts {\it Fruits} and {\it Vegetables} are represented by the unit vectors $|A_{AB}(x)\rangle=(0.73,0,0.68)$, $|B_{AB}(x)\rangle=e^{i 60.48^{\circ}}(0.69,0.55,-0.61)$ in ${\mathbb C}^{3}$. An exemplar that in Section \ref{experiment} had a big overextension with respect to {\it Pets And Farmyard Animals} was {\it Goldfish}. {\it Goldfish} scored $\mu_{x}(A)=0.925$ with respect to {\it Pets}, $\mu_{x}(B)=0.16875$ with respect to {\it Farmyard Animals} and	$\mu_{x}(A \ {\rm and} \ B)=0.425$ with respect to {\it Pets And Farmyard Animals}. It can be modeled in ${\mathbb C}^{3}\oplus ({\mathbb C}^{3}\otimes{\mathbb C}^{3})$ with $\theta_{AB}(x)=99.22^{\circ}$, $m_{AB}(x)^2=0.23$ and	$n_{AB}(x)^2=0.77$. The concept {\it Pets} is represented by $|A_{AB}(x)\rangle=(0.96,0,0.27)$, while the concept {\it Farmyard Animals} is represented by $|B_{AB}(x)\rangle=e^{i 99.22^{\circ}}(0.38,0.32,-0.91)$ (Table 7a). Another non-classical exemplar was {\it Parsley} with respect to {\it Fruits And Vegetables}. In our Fock space representation, it is possible to model $\mu_{x}(A)=0.01875$,	$\mu_{x}(B)=0.78125$ and $\mu_{x}(A \ {\rm and} \ B)=0.45$ of {\it Parsley} with $\theta_{AB}(x)=45.6^{\circ}$, $m_{AB}(x)^2=0.07$ and $n_{AB}(x)^2=0.93$. Hence, the decision process of a subject estimating whether {\it Parsley} belongs to {\it Fruits}, {\it Vegetables} and {\it Fruits And Vegetables} occurs prevalently in sector 1 of the Fock space ${\mathbb C}^{3}\oplus ({\mathbb C}^{3}\otimes{\mathbb C}^{3})$.  The concepts {\it Fruits} and {\it Vegetables} are represented by $|A_{AB}(x)\rangle=(0.99,0,0.14)$ and $|B_{AB}(x)\rangle=e^{i 45.6^{\circ}}(0.18,0.45,-0.88)$, respectively (Table 8a). But, our quantum-theoretic framework also allows the modeling of `classical data', that is, data that can be represented in a classical Kolmogorovian probability model (Section \ref{classicality}). Indeed, the exemplar {\it Shelves} had a membership weight of $\mu_{x}(A)=0.85$ with respect to {\it Home Furnishing}, $\mu_{x}(B)=0.93125$ with respect to {\it Furniture}, and $\mu_{x}(A \ {\rm and} \  B)=0.8375$ with respect to {\it Home Furnishing And Furniture}. {\it Shelves} can be represented in ${\mathbb C}^{3}\oplus ({\mathbb C}^{3}\otimes{\mathbb C}^{3})$ with $\theta_{AB}(x)=101.54^{\circ}$, $m_{AB}(x)^2=0.42$ and	$n_{AB}(x)^2=0.58$. The concept {\it Home Furnishing} is represented by	$|A_{AB}(x)\rangle=(0.92,0,0.39)$ and the concept is represented by $|B_{AB}(x)\rangle=e^{i 101.54^{\circ}}(0.37,0.96,-0.26)$ with respect to the exemplar {\it Shelves} (Table 5a).

Let us now come to the modeling of the conjunction `$A \ {\rm and} \ {\rm not} \ B$'. Tables 5b, 6b, 7b and 8b report, for each exemplar $x$, the values of the interference angle $\theta_{AB'}(x)$ and the weights $m_{AB'}(x)^2$ and $n_{AB'}(x)^2$ which satisfy Equation (\ref{AB'}), together with the representation of the unit vectors $|A_{AB'}(x)\rangle$ and $|{\rm not} \ B_{AB'}(x)\rangle$ in ${\mathbb C}^{3}$ satisfying Equations (\ref{vectorA}), (\ref{vectorB1}) and (\ref{vectorB2}). Let us start from the data that are classically  very problematical. The exemplar {\it Prize Bull} was double overextended with respect to {\it Pets And Not Farmyard Animals}, since it scored $\mu_{x}(A)=0.13125$ with respect to {\it Pets}, $\mu_{x}({\rm not} \ B)=0.2625$ with respect to {\it Not Farmyard Animals} and	$\mu_{x}(A \ {\rm and} \ {\rm not} \ B)=0.275$ wit respect to {\it Pets And Not Faryard Animals}. The exemplar {\it Prize Bull} can be modeled in Fock space with an interference angle $\theta_{AB'}(x)=45.11^{\circ}$ and weights $m_{AB'}(x)^2=0.18$ for sector 2 of Fock space and $n_{AB'}(x)^2=0.82$ for sector 1. The concepts {\it Pets} and {\it Not Farmyard Animals} are represented by	$|A_{AB'}(x)\rangle=(0.93,	0,	0.36)$	and $|{\rm not} \ B_{AB'}(x)\rangle=e^{i 45.11^{\circ}}(0.2, 0.84, -0.51)$ with respect to the exemplar {\it Prize Bull} (Table 7b). The exemplar {\it Shelves} scored a high overextension with respect to {\it Home Furnishing And Not Furniture}, since it gave $\mu_{x}(A)=0.85$, $\mu_{x}({\rm not} \ B)=0.125$ and	$\mu_{x}(A \ {\rm and} \ {\rm not} \ B)=0.3875$. It can be modeled in Fock space with an interference angle $\theta_{AB'}(x)=87.87^{\circ}$ and weights $m_{AB'}(x)^2=0.29$ and $n_{AB'}(x)^2=0.71$. The concepts {\it Home Furnishing} and {\it Not Furniture} are represented by $|A_{AB'}(x)\rangle=(0.39,	0,	0.92)$	and $|{\rm not} \ B_{AB'}(x)\rangle=e^{i 87.87^{\circ}}(0.84,	0.41,	-0.35)$ with respect to the exemplar {\it Shelves} (Table 5b). A similar pattern can be observed for the exemplar {\it Doberman Guard Dog} which scored $\mu_{x}(A)=0.88125$, $\mu_{x}({\rm not} \ B)=0.26875$ and $\mu_{x}(A \ {\rm and} \ {\rm not} \ B)=0.55$. Our quantum model works for this exemplar with  $\theta_{AB'}(x)=74.87^{\circ}$	and weights	$m_{AB'}(x)^2=0.25$ and $n_{AB'}(x)^2=0.75$. The concepts {\it Pets} and {\it Not Farmyard Animals} are represented by	$|A_{AB'}(x)\rangle=(0.94,	0,	0.34)$	and $|{\rm not} \ B_{AB'}(x)\rangle=e^{i 74.87^{\circ}}(0.31,	0.41,	-0.86)$ with respect to the exemplar {\it Doberman Guard Dog} (Table 7b). Also in this case, the `classical data' can be modeled as well. For example, the exemplar {\it Yam} scored		$\mu_{x}(A)=0.375$ with respect to {\it Fruits}, 	$\mu_{x}({\rm not} \ B)=0.43125$ with respect to {\it Not Vegetables} and $\mu_{x}(A \ {\rm and} \ {\rm not} \ B)=0.2375$ with respect to {\it Fruits And Not Vegetables}. {\it Yam} has an interference angle $\theta_{AB'}(x)=94.32^{\circ}$ and weights $m_{AB'}(x)=0.64$	and	$n_{AB'}(x)=0.36$. This means that the decision process of a subject estimating whether {\it Yam} belongs to {\it Fruits}, {\it Vegetables} and {\it Fruits And Vegetables} occurs prevalently in sector 2 of the Fock space ${\mathbb C}^{3}\oplus ({\mathbb C}^{3}\otimes{\mathbb C}^{3})$. The concept {\it Fruits} is represented by $|A_{AB'}(x)\rangle=(0.79,	0,	0.61)$, while {\it Not Vegetables} is represented by $|{\rm not} \ B_{AB'}(x)\rangle=e^{i 94.32^{\circ}}(0.51,0.56,-0.66)$ in the Hilbert space ${\mathbb C}^{3}$ (Table 8b).

Our analysis above, together with Tables 5-8, allow one to conclude that our quantum-mechanical model in Fock space satisfactorily represents the majority of experimental data collected in our experiments on concepts and their combinations, which are classically problematical, as we have observed in Section \ref{classicality}. Moreover, our quantum-theoretic framework describes the deviations of these data from classical (fuzzy set) logic and probability theory in terms of genuinely quantum effects. Indeed, a quantum probabilistic model is needed for the whole set of data, which entails the presence of `contextuality' (Aerts, 1986). Also, both `superposition' and `interference' are manifestly present between concepts, both in the conjunction  `$A \ {\rm and} \ B$' and in the conjunction `$A \ {\rm and} \ {\rm not} \ B$'. And quantum field-theoretic notions, i.e. sector, Fock space, tensor product (Section \ref{quantum}) are required to model our data. For what instead concerns `emergence', `emergent dynamics' -- which also strongly occurs -- deserves a more detailed analysis and it is connected with our explanatory hypothesis we have recently provided to cope with such deviations from classicality in cognitive and decision processes (Aerts, 2009a; Aerts, Gabora \& Sozzo, 2013). We have indeed proposed a mechanism that explains the effectiveness of a quantum-theoretic modeling. It is exactly the quantum effect of emergence which comes into play. More precisely, whenever a given subject is asked to estimate whether a given exemplar $x$ belongs to the vague concepts $A$, $B$, `$A \ {\rm and} \ B$' (`$A \ {\rm and} \ {\rm not} \ B$'), two mechanisms act simultaneously and in superposition in the subject's thought. A `quantum logical thought', which is a probabilistic version of the classical logical reasoning, where the subject considers two copies of the exemplar $x$ and estimates whether the first copy belongs to $A$ and the second copy of $x$ belongs to $B$ (`${\rm not} \ B$'). But also a `quantum conceptual thought' acts, where the subject estimates whether the exemplar $x$ belongs to the newly emergent concept `$A \ {\rm and} \ B$' (`$A \ {\rm and} \ {\rm not} \ B$'). The place whether these superposed processes can be suitably structured is the Fock space. Sector 1 of Fock space hosts the latter process, while sector 2 hosts the former, while the weights $m_{AB}^2(x)$ and $n_{AB}^2(x)$ ($m_{AB'}^2(x)$ and $n_{AB'}^2(x)$) measure the amount of `participation' of sectors 2 and 1, respectively. But, what happens in human thought during a cognitive test is a quantum superposition of both processes. As a consequence of this explanatory hypothesis, an effect, a deviation, or a contradiction, are not failures of classical logical reasoning but, rather, they are a manifestation of the presence of a superposed thought, quantum logical and quantum emergent thought.

It is important to remark, to conclude, that we did not inquire into the relationships between the representation of a concept $A$ and that of `${\rm not} \ A$'in the present article. We indeed observe that quantum logical rules should hold in Sector 2 of Fock space -- this was implicitly assumed in the modeling of both the conjunction `$A \ {\rm and} \ B$` and the disjunction `$A \ {\rm or} \ B$' in Section \ref{brussels}. Logical coherence would then lead us to assume that the representation of `${\rm not} \ A$' should be constructed from the representation of $A$ by requiring that quantum logical rules -- the rules of quantum logical negation, in this case -- are valid in Sector 2 of Fock space. We believe that this should be the case, but we also think that a complete analysis of this situation is only possible if data on $A$, $B$, `${\rm not} \ B$`, `$A \ {\rm and} \ B$', `$A \ {\rm and} \ {\rm not} \  B$', but also `${\rm not} \ A$`, `${\rm not} \ A \ {\rm and} \ B$' and `${\rm not} \ A \ {\rm and} \ {\rm not} \ B$' are simultaneously collected. We are presently working on the elaboration of these data and we plan to deal with this interesting aspect in a forthcoming paper (Aerts, Sozzo \& Veloz, 2014).

\begin{table}[scale=0.2]
\begin{center}
\footnotesize
\begin{tabular}{|cccccccll|}
\hline
\multicolumn{9}{|l|}{\it $A$=Home Furnishing, $B$=Furniture} \\
\hline					
{\it Exemplar}		&	$\mu_{x}(A)$	&	$\mu_{x}(B)$	&	$\mu_{x}(A \ {\rm and} \ B)$ &	$\theta_{AB}(x)$	&	$m_{AB}(x)^2$	&	$n_{AB}(x)^2$ & $|A_{AB}(x)\rangle$ & $e^{-i \theta_{AB}(x)}|B_{AB}(x)\rangle$	\\
\hline
{\it Mantelpiece}		&	0.9	&	0.6125	&	0.7125	&	82.84	&	0.3	&	0.7	&	(0.95,	0,	0.32)	&	(0.43,	0.75,	-0.62)	\\
{\it Window Seat}		&	0.5	&	0.48125	&	0.45	&	74.75	&	0.45	&	0.55	&	(0.71,	0,	0.71)	&	(0.76,	0.19,	-0.69)	\\
{\it Painting}		&	0.8	&	0.4875	&	0.6375	&	71.12	&	0.31	&	0.69	&	(0.89,	0,	0.45)	&	(0.52,	0.6,	-0.72)	\\
{\it Light Fixture}		&	0.875	&	0.6	&	0.725	&	73.2	&	0.28	&	0.72	&	(0.94,	0,	0.35)	&	(0.55,	0.74,	-0.63)	\\
{\it Kitchen Counter}		&	0.66875	&	0.4875	&	0.55	&	73.91	&	0.39	&	0.61	&	(0.82,	0,	0.58)	&	(0.72,	0.48,	-0.72)	\\
{\it Bath Tub}		&	0.725	&	0.5125	&	0.5875	&	74.9	&	0.37	&	0.63	&	(0.85,	0,	0.52)	&	(0.68,	0.57,	-0.7)	\\
{\it Deck Chair}		&	0.73125	&	0.9	&	0.7375	&	98.46	&	0.4	&	0.6	&	(0.86,	0,	0.52)	&	(0.39,	0.93,	-0.32)	\\
{\it Shelves}		&	0.85	&	0.93125	&	0.8375	&	101.54	&	0.42	&	0.58	&	(0.92,	0,	0.39)	&	(0.37,	0.96,	-0.26)	\\
{\it Rug}		&	0.89375	&	0.575	&	0.7	&	79.31	&	0.28	&	0.72	&	(0.95,	0,	0.33)	&	(0.46,	0.72,	-0.65)	\\
{\it Bed}		&	0.75625	&	0.925	&	0.7875	&	93.14	&	0.34	&	0.66	&	(0.87,	0,	0.49)	&	(0.36,	0.95,	-0.27)	\\
{\it Wall-Hangings}		&	0.86875	&	0.4625	&	0.55	&	95.81	&	0.37	&	0.63	&	(0.93,	0,	0.36)	&	(0.55,	0.62,	-0.73)	\\
{\it Space Rack}		&	0.375	&	0.425	&	0.4125	&	68.21	&	0.35	&	0.65	&	(0.79,	0,	0.61)	&	(0.53,	0.57,	-0.65)	\\
{\it Ashtray}		&	0.74375	&	0.4	&	0.4875	&	82.43	&	0.42	&	0.58	&	(0.86,	0,	0.51)	&	(0.63,	0.44,	-0.77)	\\
{\it Bar}		&	0.71875	&	0.625	&	0.6125	&	80.53	&	0.41	&	0.59	&	(0.85,	0,	0.53)	&	(0.51,	0.69,	-0.61)	\\
{\it Lamp}		&	0.94375	&	0.64375	&	0.75	&	88	&	0.25	&	0.75	&	(0.97,	0,	0.24)	&	(0.32,	0.79,	-0.6)	\\
{\it Wall Mirror}		&	0.9125	&	0.75625	&	0.825	&	73.68	&	0.27	&	0.73	&	(0.96,	0,	0.3)	&	(0.39,	0.86,	-0.49)	\\
{\it Door Bell}		&	0.75	&	0.33125	&	0.5	&	75.7	&	0.36	&	0.64	&	(0.87,	0,	0.5)	&	(0.57,	0.33,	-0.82)	\\
{\it Hammock}		&	0.61875	&	0.6625	&	0.6	&	76.5	&	0.4	&	0.6	&	(0.79,	0,	0.62)	&	(0.64,	0.67,	-0.58)	\\
{\it Desk}		&	0.78125	&	0.95	&	0.775	&	130.06	&	0.42	&	0.58	&	(0.88,	0,	0.47)	&	(0.27,	0.97,	-0.22)	\\
{\it Refrigerator}		&	0.74375	&	0.725	&	0.6625	&	85.71	&	0.43	&	0.57	&	(0.86,	0,	0.51)	&	(0.53,	0.79,	-0.52)	\\
{\it Park Bench}		&	0.53125	&	0.6625	&	0.55	&	76.77	&	0.41	&	0.59	&	(0.73,	0,	0.68)	&	(0.64,	0.6,	-0.58)	\\
{\it Waste Paper Basket}		&	0.69375	&	0.54375	&	0.5875	&	74.8	&	0.38	&	0.62	&	(0.83,	0,	0.55)	&	(0.54,	0.59,	-0.68)	\\
{\it Sculpture}		&	0.825	&	0.4625	&	0.575	&	82.95	&	0.35	&	0.65	&	(0.91,	0,	0.42)	&	(0.51,	0.59,	-0.73)	\\
{\it Sink Unit}		&	0.70625	&	0.56875	&	0.6	&	76.05	&	0.38	&	0.62	&	(0.84,	0,	0.54)	&	(0.58,	0.62,	-0.66)	\\
\hline
\end{tabular}
\normalsize
\end{center}
{\rm Table 5a.} Representation of $A$, $B$ and `$A \ {\rm and} \ B$' in the case of the concepts {\it Home Furnishing} and {\it Furniture}. Note that the angles are expressed in degrees.
\end{table}

\begin{table}[scale=0.2]
\begin{center}
\scriptsize
\begin{tabular}{|cccccccll|}
\hline
\multicolumn{9}{|l|}{\it $A$=Home Furnishing, $B$=Furniture} \\
\hline
{\it Exemplar}		&	$\mu_{x}(A)$	&	$\mu_{x}({\rm not} \ B)$	&	$\mu_{x}(A \ {\rm and} \ {\rm not} \ B)$ &	$\theta_{AB'}(x)$	&	$m_{AB'}(x)^2$	&	$n_{AB'}(x)^2$ & $|A_{AB'}(x)\rangle$ & $e^{-i \theta_{AB'}(x)}|{\rm not} \ B_{AB'}(x)\rangle$	\\
\hline
{\it Mantelpiece}		&	0.9	&	0.5	&	0.75	&	57.08	&	0.19	&	0.81	&	(0.95,	0,	0.32)	&	(0.24,	0.67,	-0.71)	\\
{\it Window Seat}		&	0.5	&	0.55	&	0.4875	&	74.53	&	0.44	&	0.56	&	(0.71,	0,	0.71)	&	(0.67,	0.32,	-0.67)	\\
{\it Painting}		&	0.8	&	0.64375	&	0.6	&	97.17	&	0.51	&	0.49	&	(0.89,	0,	0.45)	&	(0.3,	0.74,	-0.6)	\\
{\it Light Fixture}		&	0.875	&	0.5125	&	0.625	&	86.17	&	0.33	&	0.67	&	(0.94,	0,	0.35)	&	(0.26,	0.67,	-0.7)	\\
{\it Kitchen Counter}		&	0.66875	&	0.61875	&	0.5375	&	87.4	&	0.5	&	0.5	&	(0.82,	0,	0.58)	&	(0.43,	0.66,	-0.62)	\\
{\it Bath Tub}		&	0.725	&	0.4625	&	0.5875	&	70.93	&	0.34	&	0.66	&	(0.85,	0,	0.52)	&	(0.45,	0.51,	-0.73)	\\
{\it Deck Chair}		&	0.73125	&	0.2	&	0.4125	&	77.99	&	0.33	&	0.67	&	(0.52,	0,	0.86)	&	(0.74,	0.51,	-0.45)	\\
{\it Shelves}		&	0.85	&	0.125	&	0.3875	&	87.87	&	0.29	&	0.71	&	(0.39,	0,	0.92)	&	(0.84,	0.41,	-0.35)	\\
{\it Rug}		&	0.89375	&	0.60625	&	0.675	&	91.51	&	0.34	&	0.66	&	(0.95,	0,	0.33)	&	(0.22,	0.75,	-0.63)	\\
{\it Bed}		&	0.75625	&	0.10625	&	0.3625	&	84.04	&	0.26	&	0.74	&	(0.49,	0,	0.87)	&	(0.57,	0.75,	-0.33)	\\
{\it Wall-Hangings}		&	0.86875	&	0.68125	&	0.7125	&	87.79	&	0.37	&	0.63	&	(0.93,	0,	0.36)	&	(0.22,	0.8,	-0.56)	\\
{\it Space Rack}		&	0.375	&	0.61875	&	0.4875	&	71.12	&	0.39	&	0.61	&	(0.79,	0,	0.61)	&	(0.61,	0.1,	-0.79)	\\
{\it Ashtray}		&	0.74375	&	0.6375	&	0.6	&	87.3	&	0.45	&	0.55	&	(0.86,	0,	0.51)	&	(0.35,	0.72,	-0.6)	\\
{\it Bar}		&	0.71875	&	0.50625	&	0.6125	&	70	&	0.34	&	0.66	&	(0.85,	0,	0.53)	&	(0.44,	0.56,	-0.7)	\\
{\it Lamp}		&	0.94375	&	0.4875	&	0.7	&	75.28	&	0.2	&	0.8	&	(0.97,	0,	0.24)	&	(0.17,	0.68,	-0.72)	\\
{\it Wall Mirror}		&	0.9125	&	0.45	&	0.6625	&	74.9	&	0.23	&	0.77	&	(0.96,	0,	0.3)	&	(0.23,	0.63,	-0.74)	\\
{\it Door Bell}		&	0.75	&	0.7875	&	0.6375	&	104.71	&	0.61	&	0.39	&	(0.87,	0,	0.5)	&	(0.27,	0.85,	-0.46)	\\
{\it Hammock}		&	0.61875	&	0.40625	&	0.5	&	71.51	&	0.4	&	0.6	&	(0.79,	0,	0.62)	&	(0.6,	0.2,	-0.77)	\\
{\it Desk}		&	0.78125	&	0.0875	&	0.325	&	94.73	&	0.25	&	0.75	&	(0.47,	0, 0.88)	&	(0.56,	0.77,	-0.3)	\\
{\it Refrigerator}		&	0.74375	&	0.40625	&	0.55	&	73.67	&	0.35	&	0.65	&	(0.86,	0,	0.51)	&	(0.45,	0.45,	-0.77)	\\
{\it Park Bench}		&	0.53125	&	0.45625	&	0.2875	&	94.77	&	0.79	&	0.21	&	(0.68,	0,	0.73)	&	(0.72,	0.16,	-0.68)	\\
{\it Waste Paper Basket}		&	0.69375	&	0.63125	&	0.4125	&	118.07	&	0	&	1	&	(0.83,	0,	0.55)	&	(0.4,	0.68,	-0.61)	\\
{\it Sculpture}		&	0.825	&	0.65625	&	0.725	&	73.65	&	0.32	&	0.68	&	(0.91,	0,	0.42)	&	(0.27,	0.76,	-0.59)	\\
{\it Sink Unit}		&	0.70625	&	0.575	&	0.5625	&	82.77	&	0.44	&	0.56	&	(0.84,	0,	0.54)	&	(0.42,	0.63,	-0.65)	\\
\hline
\end{tabular}
\normalsize
\end{center}
{\rm Table 5b.} Representation of $A$, `${\rm not} \ B$' and `$A \  {\rm and} \ {\rm not} \ B$' in the case of the concepts {\it Home Furnishing} and {\it Furniture}. Note that the angles are expressed in degrees.
\end{table}

\begin{table}[\scale=0.02]
\begin{center}
\footnotesize
\begin{tabular}{|cccccccll|}
\hline
\multicolumn{9}{|l|}{\it $A$=Spices, $B$=Herbs} \\
\hline					
{\it Exemplar}		&	$\mu_{x}(A)$	&	$\mu_{x}(B)$	&	$\mu_{x}(A \ {\rm and} \ B)$ &	$\theta_{AB}(x)$	&	$m_{AB}(x)^2$	&	$n_{AB}(x)^2$ & $|A_{AB}(x)\rangle$ & $e^{-i \theta_{AB}(x)}|B_{AB}(x)\rangle$	\\
\hline
{\it Molasses}		&	0.3625	&	0.13125	&	0.2375	&	72.46	&	0.28	&	0.72	&	(0.8,	0,	0.6)	&	(0.26,	0.89,	-0.36)	\\
{\it Salt}		&	0.66875	&	0.04375	&	0.2375	&	113.97	&	0.19	&	0.81	&	(0.58,	0,	0.82)	&	(0.22,	0.93,	-0.21)	\\
{\it Peppermint}		&	0.66875	&	0.925	&	0.7	&	107.92	&	0.37	&	0.63	&	(0.82,	0,	0.58)	&	(0.41,	0.94,	-0.27)	\\
{\it Curry}		&	0.9625	&	0.28125	&	0.5375	&	100.93	&	0.17	&	0.83	&	(0.98,	0,	0.19)	&	(0.36,	0.5,	-0.85)	\\
{\it Oregano}		&	0.8125	&	0.85625	&	0.7875	&	86.59	&	0.38	&	0.62	&	(0.9,	0,	0.43)	&	(0.6,	0.91,	-0.38)	\\
{\it MSG}		&	0.44375	&	0.11875	&	0.225	&	84.18	&	0.32	&	0.68	&	(0.75,	0,	0.67)	&	(0.27,	0.89,	-0.34)	\\
{\it Chili Pepper}		&	0.975	&	0.53125	&	0.8	&	44.34	&	0.1	&	0.9	&	(0.99,	0,	0.16)	&	(0.31,	0.72,	-0.68)	\\
{\it Mustard}		&	0.65	&	0.275	&	0.4875	&	66.61	&	0.32	&	0.68	&	(0.59,	0,	0.81)	&	(0.62,	0.46,	-0.52)	\\
{\it Mint}		&	0.64375	&	0.95625	&	0.7875	&	75.75	&	0.2	&	0.8	&	(0.8,	0,	0.6)	&	(0.37,	0.97,	-0.21)	\\
{\it Cinnamon}		&	1	&	0.49375	&	0.6875	&	0	&	0.23	&	0.77	&	(1,	0,	0)	&	(0,	0.7,	-0.71)	\\
{\it Parsley}		&	0.5375	&	0.9	&	0.675	&	81.74	&	0.28	&	0.72	&	(0.73,	0,	0.68)	&	(0.51,	0.9,	-0.32)	\\
{\it Saccarin}		&	0.34375	&	0.1375	&	0.2375	&	70.82	&	0.28	&	0.72	&	(0.81,	0,	0.59)	&	(0.24,	0.89, -0.37)	\\
{\it Poppy Seeds}		&	0.81875	&	0.46875	&	0.5875	&	80.44	&	0.35	&	0.65	&	(0.9,	0,	0.43)	&	(0.59,	0.59,	-0.73)	\\
{\it Pepper}		&	0.99375	&	0.46875	&	0.7	&	102.84	&	0.07	&	0.93	&	(1,	0,	0.08)	&	(0.16,	0.68,	-0.73)	\\
{\it Turmeric}		&	0.88125	&	0.525	&	0.7375	&	61.68	&	0.22	&	0.78	&	(0.94,	0,	0.34)	&	(0.52,	0.68,	-0.69)	\\
{\it Sugar}		&	0.45	&	0.34375	&	0.35	&	76.84	&	0.41	&	0.59	&	(0.74,	0,	0.67)	&	(0.49,	0.61,	-0.59)	\\
{\it Vinegar} 		&	0.3	&	0.10625	&	0.15	&	87.31	&	0.34	&	0.66	&	(0.84,	0,	0.55)	&	(0.2,	0.92,	-0.33)	\\
{\it Sesame Seeds}		&	0.8	&	0.4875	&	0.5875	&	80.12	&	0.36	&	0.64	&	(0.89,	0,	0.45)	&	(0.6,	0.6,	-0.72)	\\
{\it Lemon Juice}		&	0.275	&	0.2	&	0.15	&	91.91	&	0.46	&	0.54	&	(0.85,	0,	0.52)	&	(0.26,	0.85,	-0.45)	\\
{\it Chocolate}		&	0.26875	&	0.2125	&	0.2	&	79.79	&	0.37	&	0.63	&	(0.86,	0,	0.52)	&	(0.27,	0.84,	-0.46)	\\
{\it Horseradish}		&	0.6125	&	0.66875	&	0.6125	&	74.5	&	0.38	&	0.62	&	(0.78,	0,	0.62)	&	(0.63,	0.68,	-0.58)	\\
{\it Vanilla}		&	0.7625	&	0.5125	&	0.625	&	72.11	&	0.33	&	0.67	&	(0.87,	0,	0.49)	&	(0.58,	0.6,	-0.7)	\\
{\it Chives}		&	0.6625	&	0.8875	&	0.7625	&	73.68	&	0.28	&	0.72	&	(0.81,	0,	0.58)	&	(0.35,	0.91,	-0.34)	\\
{\it Root Ginger}		&	0.84375	&	0.5625	&	0.6875	&	73.43	&	0.3	&	0.7	&	(0.92,	0,	0.4)	&	(0.55,	0.69,	-0.66)	\\
\hline
\end{tabular}
\normalsize
\end{center}
{\rm Table 6a.} Representation of $A$, $B$ and `$A \ {\rm and} \ B$' in the case of the concepts {\it Spices} and {\it Herbs}. Note that the angles are expressed in degrees.
\end{table}

\begin{table}[scale=0.2]
\begin{center}
\scriptsize
\begin{tabular}{|cccccccll|}
\hline
\multicolumn{9}{|l|}{\it $A$=Spices, $B$=Herbs} \\
\hline
{\it Exemplar}		&	$\mu_{x}(A)$	&	$\mu_{x}({\rm not} \ B)$	&	$\mu_{x}(A \ {\rm and} \ {\rm not} \ B)$ &	$\theta_{AB'}(x)$	&	$m_{AB'}(x)^2$	&	$n_{AB'}(x)^2$ & $|A_{AB'}(x)\rangle$ & $e^{-i \theta_{AB'}(x)}|{\rm not} \ B_{AB'}(x)\rangle$	\\
\hline
{\it Molasses}		&	0.3625	&	0.8375	&	0.5375	&	81.2	&	0.32	&	0.68	&	(0.6,	0,	0.8)	&	(0.53,	0.74,	-0.4)	\\
{\it Salt}		&	0.66875	&	0.91875	&	0.6875	&	110.36	&	0.4	&	0.6	&	(0.82,	0,	0.58)	&	(0.2,	0.94,	-0.29)	\\
{\it Peppermint}		&	0.66875	&	0.1	&	0.375	&	72.08	&	0.22	&	0.78	&	(0.58,	0,	0.82)	&	(0.45,	0.84,	-0.32)	\\
{\it Curry}		&	0.9625	&	0.775	&	0.875	&	66.1	&	0.19	&	0.81	&	(0.98,	0,	0.19)	&	(0.09,	0.88,	-0.47)	\\
{\it Oregano}		&	0.8125	&	0.125	&	0.4	&	82.46	&	0.27	&	0.73	&	(0.43,	0,	0.9)	&	(0.74,	0.58,	-0.35)	\\
{\it MSG}		&	0.44375	&	0.85	&	0.575	&	84.41	&	0.34	&	0.66	&	(0.67,	0,	0.75)	&	(0.43,	0.81,	-0.39)	\\
{\it Chili Pepper}		&	0.975	&	0.5625	&	0.9	&	0	&	0	&	1.03	&	(0.99,	0,	0.16)	&	(0.11,	0.74,	-0.66)	\\
{\it Mustard}		&	0.65	&	0.70625	&	0.65	&	75.03	&	0.37	&	0.63	&	(0.81,	0,	0.59)	&	(0.4,	0.74,	-0.54)	\\
{\it Mint}		&	0.64375	&	0.0875	&	0.3125	&	82.93	&	0.24	&	0.76	&	(0.6,	0,	0.8)	&	(0.4,	0.87,	-0.3)	\\
{\it Cinnamon}		&	1	&	0.5125	&	0.7875	&	8.65	&	0	&	1	&	(1,	0,	0)	&	(0,	0.72,	-0.7)	\\
{\it Parsley}		&	0.5375	&	0.0875	&	0.2625	&	83.33	&	0.26	&	0.74	&	(0.68,	0,	0.73)	&	(0.32,	0.9,	-0.3)	\\
{\it Saccarin}		&	0.34375	&	0.875	&	0.5375	&	84.53	&	0.3	&	0.7	&	(0.59,	0,	0.81)	&	(0.49,	0.8,	-0.35)	\\
{\it Poppy Seeds}		&	0.81875	&	0.5375	&	0.6625	&	73.09	&	0.31	&	0.69	&	(0.9,	0,	0.43)	&	(0.32,	0.66,	-0.68)	\\
{\it Pepper}		&	0.99375	&	0.58125	&	0.9	&	0	&	0	&	1	&	(1,	0,	0.08)	&	(0.05,	0.76,	-0.65)	\\
{\it Turmeric}		&	0.88125	&	0.43125	&	0.6875	&	63.09	&	0.22	&	0.78	& (0.94,	0,	0.34)	&	(0.28,	0.6,	-0.75)	\\
{\it Sugar}		&	0.45	&	0.76875	&	0.5625	&	77.55	&	0.36	&	0.64	&	(0.67,	0,	0.74)	&	(0.53,	0.7,	-0.48)	\\
{\it Vinegar}		&	0.3	&	0.88125	&	0.4125	&	108.16	&	0.37	&	0.63	&	(0.55,	0,	0.84)	&	(0.53,	0.78,	-0.34)	\\
{\it Sesame Seeds}		&	0.8	&	0.5875	&	0.7	&	68.75	&	0.3	&	0.7	&	(0.89,	0,	0.45)	&	(0.32,	0.7,	-0.64)	\\
{\it Lemon Juice}		&	0.275	&	0.80625	&	0.425	&	87.97	&	0.39	&	0.61	&	(0.52,	0,	0.85) &	(0.71,	0.54,	-0.44)	\\
{\it Chocolate}		&	0.26875	&	0.8	&	0.4625	&	80.83	&	0.35	&	0.65	&	(0.52,	0,	0.86)	&	(0.74,	0.51,	-0.45)	\\
{\it Horseradish}		&	0.6125	&	0.28125	&	0.4	&	76.48	&	0.39	&	0.61	&	(0.62,	0,	0.78)	&	(0.67,	0.52,	-0.53)	\\
{\it Vanilla}		&	0.7625	&	0.4875	&	0.6125	&	72.05	&	0.33	&	0.67	&	(0.87,	0,	0.49)	&	(0.4,	0.57,	-0.72)	\\
{\it Chives}		&	0.6625	&	0.25625	&	0.275	&	96.58	&	0.57	&	0.43	&	(0.58,	0,	0.81)	&	(0.71,	0.49,	-0.51)	\\
{\it Root Ginger}		&	0.84375	&	0.44375	&	0.5875	&	81	&	0.32	&	0.68	&	(0.92,	0,	0.4)	&	(0.32,	0.58,	-0.75)	\\
\hline
\end{tabular}
\normalsize
\end{center}
{\rm Table 6b.} Representation of $A$, `${\rm not} \ B$' and `$A \  {\rm and} \ {\rm not} \ B$' in the case of the concepts {\it Spices} and {\it Herbs}. Note that the angles are expressed in degrees.
\end{table}

\begin{table}[scale=0.2]
\begin{center}
\footnotesize
\begin{tabular}{|cccccccll|}
\hline
\multicolumn{9}{|l|}{\it $A$=Pets, $B$=Farmyard Animals} \\
\hline					
{\it Exemplar}		&	$\mu_{x}(A)$	&	$\mu_{x}(B)$	&	$\mu_{x}(A \ {\rm and} \ B)$ &	$\theta_{AB}(x)$	&	$m_{AB}(x)^2$	&	$n_{AB}(x)^2$ & $|A_{AB}(x)\rangle$ & $e^{-i \theta_{AB}(x)}|B_{AB}(x)\rangle$	\\
\hline
{\it Goldfish}		&	0.925	&	0.16875	&	0.425	&	99.22	&	0.23	&	0.77	&	(0.96,	0,	0.27)	&	(0.38,	0.32,	-0.91)	\\
{\it Robin}		&	0.275	&	0.3625	&	0.3125	&	71.13	&	0.34	&	0.66	&	(0.85,	0,	0.52)	&	(0.46,	0.71,	-0.6)	\\
{\it Blue-tit}		&	0.25	&	0.3125	&	0.175	&	92.34	&	0.49	&	0.51	&	(0.87,	0,	0.5)	&	(0.37,	0.76,	-0.56)	\\
{\it Collie Dog}		&	0.95	&	0.76875	&	0.8625	&	68.33	&	0.22	&	0.78	&	(0.97,	0,	0.22)	&	(0.32,	0.87,	-0.48)	\\
{\it Camel}		&	0.15625	&	0.25625	&	0.2	&	71.79	&	0.3	&	0.7	&	(0.92,	0,	0.4)	&	(0.24,	0.83,	-0.51)	\\
{\it Squirrel}		&	0.3	&	0.39375	&	0.275	&	82.07	&	0.43	&	0.57	&	(0.84,	0,	0.55)	&	(0.45,	0.66,	-0.63)	\\
{\it Guide Dog for Blind}		&	0.925	&	0.325	&	0.55	&	89.47	&	0.24	&	0.76	&	(0.96,	0,	0.27)	&	(0.39,	0.52,	-0.82)	\\
{\it Spider}		&	0.3125	&	0.3875	&	0.3125	&	76.19	&	0.39	&	0.61	&	(0.83,	0,	0.56)	&	(0.49,	0.66,	-0.62)	\\
{\it Homing Pigeon}		&	0.40625	&	0.70625	&	0.5625	&	69.14	&	0.34	&	0.66	&	(0.64,	0,	0.77)	&	(0.72,	0.53,	-0.54)	\\
{\it Monkey}		&	0.39375	&	0.175	&	0.2	&	88.75	&	0.41	&	0.59	&	(0.78,	0,	0.63)	&	(0.34,	0.84,	-0.42)	\\
{\it Circus Horse}		&	0.3	&	0.48125	&	0.3375	&	78.04	&	0.41	&	0.59	&	(0.84,	0,	0.55)	&	(0.55,	0.56,	-0.69)	\\
{\it Prize Bull}		&	0.13125	&	0.7625	&	0.425	&	73.97	&	0.25	&	0.75	&	(0.93,	0,	0.36)	&	(0.54,	0.35,	-0.87)	\\
{\it Rat}		&	0.2	&	0.35625	&	0.2125	&	84.23	&	0.4	&	0.6	&	(0.89,	0,	0.45)	&	(0.34,	0.74,	-0.6)	\\
{\it Badger}		&	0.1625	&	0.275	&	0.1375	&	92.6	&	0.44	&	0.56	&	(0.92,	0,	0.4)	&	(0.26,	0.82,	-0.52)	\\
{\it Siamese Cat}		&	0.9875	&	0.5	&	0.7375	&	74.53	&	0.1	&	0.9	&	(0.99,	0,	0.11)	&	(0.16,	0.7,	-0.71)	\\
{\it Race Horse}		&	0.2875	&	0.7	&	0.5125	&	67.6	&	0.33	&	0.67	&	(0.84,	0,	0.54)	&	(0.8,	0.13,	-0.84)	\\
{\it Fox}		&	0.13125	&	0.3	&	0.175	&	81.81	&	0.34	&	0.66	&	(0.93,	0,	0.36)	&	(0.26,	0.81,	-0.55)	\\
{\it Donkey}		&	0.2875	&	0.9	&	0.5625	&	76.72	&	0.23	&	0.77	&	(0.54,	0,	0.84)	& (0.56,	0.81,	-0.32)	\\
{\it Field Mouse}		&	0.1625	&	0.40625	&	0.225	&	83.36	&	0.36	&	0.64	&	(0.92,	0,	0.4)	&	(0.34,	0.72,	-0.64)	\\
{\it Ginger Tom-cat}		&	0.81875	&	0.50625	&	0.5875	&	84.52	&	0.37	&	0.63	&	(0.9,	0,	0.43)	&	(0.56,	0.63,	-0.7)	\\
{\it Husky in Slead team}		&	0.64375	&	0.50625	&	0.5625	&	71.71	&	0.38	&	0.62	&	(0.8,	0,	0.6)	&	(0.78,	0.48,	-0.7)	\\
{\it Cart Horse}		&	0.26875	&	0.8625	&	0.525	&	77.36	&	0.27	&	0.73	&	(0.52,	0,	0.86)	&	(0.67,	0.7,	-0.37)	\\
{\it Chicken}		&	0.23125	&	0.95	&	0.575	&	74.57	&	0.16	&	0.84	&	(0.48,	0,	0.88)	&	(0.47,	0.89,	-0.22)	\\
{\it Doberman Guard Dog}		&	0.88125	&	0.75625	&	0.8	&	76.33	&	0.31	&	0.69	&	(0.94,	0,	0.34)	&	(0.36,	0.85,	-0.49)	\\
\hline
\end{tabular}
\normalsize
\end{center}
{\rm Table 7a.} Representation of $A$, $B$ and `$A \ {\rm and} \ B$' in the case of the concepts {\it Pets} and {\it Farmyard Animals}. Note that the angles are expressed in degrees.
\end{table}

\begin{table}[scale=0.2]
\begin{center}
\scriptsize
\begin{tabular}{|cccccccll|}
\hline
\multicolumn{9}{|l|}{\it $A$=Pets, $B$=Farmyard Animals} \\
\hline					
{\it Exemplar}		&	$\mu_{x}(A)$	&	$\mu_{x}(B)$	&	$\mu_{x}(A \ {\rm and} \ B)$ &	$\theta_{AB}(x)$	&	$m_{AB}(x)^2$	&	$n_{AB}(x)^2$ & $|A_{AB}(x)\rangle$ & $e^{-i \theta_{AB}(x)}|B_{AB}(x)\rangle$	\\
\hline
{\it Goldfish}		&	0.925	&	0.8125	&	0.9125	&	48.35	&	0.18	&	0.82	&	(0.96,	0,	0.27)	&	(0.12,	0.89,	-0.43)	\\
{\it Robin}		&	0.275	&	0.6375	&	0.35	&	84.7	&	0.45	&	0.55	&	(0.85,	0,	0.52)	&	(0.49,	0.35,	-0.8)	\\
{\it Blue-tit}		&	0.25	&	0.7125	&	0.3875	&	82.83	&	0.41	&	0.59	&	(0.87,	0,	0.5)	&	(0.49,	0.22,	-0.84)	\\
{\it Collie Dog}		&	0.95	&	0.35	&	0.5625	&	99.04	&	0.2	&	0.8	&	(0.97,	0,	0.22)	&	(0.18,	0.56,	-0.81)	\\
{\it Camel}		&	0.15625	&	0.75	&	0.3125	&	94.25	&	0.37	&	0.63	&	(0.92,	0,	0.4)	&	(0.37,	0.33,	-0.87)	\\
{\it Squirrel}		&	0.3	&	0.65	&	0.2625	&	98.76	&	0.68	&	0.32	&	(0.84,	0,	0.55)	&	(0.53,	0.27,	-0.81)	\\
{\it Guide Dog for Blind}		&	0.925	&	0.69375	&	0.725	&	103.83	&	0.37	&	0.63	&	(0.96,	0,	0.27)	&	(0.16,	0.82,	-0.55)	\\
{\it Spider}		&	0.3125	&	0.63125	&	0.3125	&	91.03	&	0.57	&	0.43	&	(0.83,	0,	0.56)	&	(0.54,	0.29,	-0.79)	\\
{\it Homing Pigeon}		&	0.40625	&	0.3375	&	0.25	&	89.22	&	0.53	&	0.47	&	(0.77,	0,	0.64)	&	(0.48,	0.66,	-0.58)	\\
{\it Monkey}		&	0.39375	&	0.79375	&	0.4875	&	87.49	&	0.41	&	0.59	&	(0.63,	0,	0.78)	&	(0.56,	0.69,	-0.45)	\\
{\it Circus Horse}		&	0.3	&	0.6	&	0.35	&	83.63	&	0.46	&	0.54	&	(0.84,	0,	0.55)	&	(0.51,	0.38,	-0.77)	\\
{\it Prize Bull}		&	0.13125	&	0.2625	&	0.275	&	45.11	&	0.18	&	0.82	&	(0.93,	0,	0.36)	&	(0.2,	0.84,	-0.51)	\\
{\it Rat}		&	0.2	&	0.675	&	0.275	&	96.25	&	0.47	&	0.53	&	(0.89,	0,	0.45)	&	(0.41,	0.4,	-0.82)	\\
{\it Badger}		&	0.1625	&	0.73125	&	0.2625	&	102.33	&	0.44	&	0.56	&	(0.92,	0,	0.4)	&	(0.38,	0.36,	-0.86)	\\
{\it Siamese Cat}		&	0.9875	&	0.525	&	0.75	&	74.65	&	0.1	&	0.9	&	(0.99,	0,	0.11)	&	(0.08,	0.72,	-0.69)	\\
{\it Race Horse}		&	0.2875	&	0.3875	&	0.3125	&	74.3	&	0.36	&	0.64	&	(0.84,	0,	0.54)	&	(0.4,	0.68,	-0.62)	\\
{\it Fox}		&	0.13125	&	0.68125	&	0.2875	&	93.4	&	0.34	&	0.66	&	(0.93,	0,	0.36)	&	(0.32,	0.46, -0.83)	\\
{\it Donkey}		&	0.2875	&	0.15	&	0.175	&	82.16	&	0.35	&	0.65	&	(0.84,	0,	0.54)	&	(0.25,	0.89,	-0.39)	\\
{\it Field Mouse}		&	0.1625	&	0.5875	&	0.2375	&	96.42	&	0.42	&	0.58	&	(0.92,	0, 0.4)	&	(0.34,	0.55,	-0.77)	\\
{\it Ginger Tom-cat}		&	0.81875	&	0.54375	&	0.575	&	91.68	&	0.43	&	0.57	&	(0.9,	0,	0.43)	&	(0.32,	0.67,	-0.68)	\\
{\it Husky in Slead Team}		&	0.64375	&	0.525	&	0.5125	&	80.06	&	0.45	&	0.55	&	(0.8,	0,	0.6)	&	(0.51,	0.51,	-0.69)	\\
{\it Cart Horse}		&	0.26875	&	0.15	&	0.2	&	72.68	&	0.3	&	0.7	&	(0.86,	0,	0.52)	&	(0.23,	0.89,	-0.39)	\\
{\it Chicken}		&	0.23125	&	0.0625	&	0.1125	&	86.61	&	0.3	&	0.7	&	(0.88,	0,	0.48)	&	(0.14,	0.96,	-0.25)	\\
{\it Doberman Guard Dog}		&	0.88125	&	0.26875	&	0.55	&	74.87	&	0.25	&	0.75	&	(0.94,	0,	0.34)	&	(0.31,	0.41,	-0.86)	\\
\hline
\end{tabular}
\normalsize
\end{center}
{\rm Table 7b.} Representation of $A$, `${\rm not} \ B$' and `$A \  {\rm and} \ {\rm not} \ B$' in the case of the concepts {\it Pets} and {\it Farmyard Animals}. Note that the angles are expressed in degrees.
\end{table}

\begin{table}[scale=0.2]
\begin{center}
\footnotesize
\begin{tabular}{|cccccccll|}
\hline
\multicolumn{9}{|l|}{\it $A$=Fruits, $B$=Vegetables} \\
\hline					
{\it Exemplar}		&	$\mu_{x}(A)$	&	$\mu_{x}(B)$	&	$\mu_{x}(A \ {\rm and} \ B)$ &	$\theta_{AB}(x)$	&	$m_{AB}(x)^2$	&	$n_{AB}(x)^2$ & $|A_{AB}(x)\rangle$ & $e^{-i \theta_{AB}(x)}|B_{AB}(x)\rangle$	\\
\hline
{\it Apple}		&	1	&	0.225	&	0.6	&	0	&	0.03	&	0.97	&	(1,	0,	0)	&	(0,	0.47,	-0.88)	\\
{\it Parsley}		&	0.01875	&	0.78125	&	0.45	&	45.6	&	0.07	&	0.93	&	(0.99,	0,	0.14)	&	(0.18,	0.45,	-0.88)	\\
{\it Olive}		&	0.53125	&	0.63125	&	0.65	&	60.48	&	0.3	&	0.7	&	(0.73,	0,	0.68)	&	(0.69,	0.55,	-0.61)	\\
{\it Chili Pepper}		&	0.1875	&	0.73125	&	0.5125	&	61.75	&	0.25	&	0.75	&	(0.9,	0,	0.43)	&	(0.56,	0.32,	-0.86)	\\
{\it Broccoli}		&	0.09375	&	1	&	0.5875	&	0	&	-0.09	&	1.09	&	(0.31,	0,	0.95)	&	(0,	1,	0)	\\
{\it Root Ginger}		&	0.1375	&	0.7125	&	0.4625	&	63.12	&	0.22	&	0.78	&	(0.93,	0,	0.37)	&	(0.48,	0.42,	-0.84)	\\
{\it Pumpkin}		&	0.45	&	0.775	&	0.6625	&	61.83	&	0.27	&	0.73	&	(0.67,	0,	0.74)	&	(0.84,	0.71,	-0.47)	\\
{\it Raisin}		&	0.88125	&	0.26875	&	0.525	&	79.77	&	0.26	&	0.74	&	(0.94,	0,	0.34)	&	(0.51,	0.41,	-0.86)	\\
{\it Acorn}		&	0.5875	&	0.4	&	0.4625	&	73.7	&	0.42	&	0.58	&	(0.64,	0,	0.77)	&	(0.68,	0.17,	-0.63)	\\
{\it Mustard}		&	0.06875	&	0.3875	&	0.2875	&	48.67	&	0.16	&	0.84	&	(0.97,	0,	0.26)	&	(0.19,	0.76,	-0.62)	\\
{\it Rice}		&	0.11875	&	0.45625	&	0.2125	&	88.8	&	0.34	&	0.66	&	(0.94,	0,	0.34)	&	(0.3,	0.69,	-0.68)	\\
{\it Tomato}		&	0.3375	&	0.8875	&	0.7	&	51.31	&	0.17	&	0.83	&	(0.58,	0,	0.81)	&	(0.58,	0.82,	-0.34)	\\
{\it Coconut}		&	0.925	&	0.31875	&	0.5625	&	85.23	&	0.23	&	0.77	&	(0.96,	0,	0.27)	&	(0.39,	0.51,	-0.83)	\\
{\it Mushroom}		&	0.11875	&	0.6625	&	0.325	&	83.53	&	0.28	&	0.72	&	(0.94,	0,	0.34)	&	(0.4,	0.5,	-0.81)	\\
{\it Wheat}		&	0.16875	&	0.50625	&	0.3375	&	70	&	0.28	&	0.72	&	(0.91,	0,	0.41)	&	(0.39,	0.63,	-0.71)	\\
{\it Green Pepper}		&	0.225	&	0.6125	&	0.4875	&	59.33	&	0.26	&	0.74	&	(0.88,	0,	0.47)	&	(0.57,	0.46,	-0.78)	\\
{\it Watercress}		&	0.1375	&	0.7625	&	0.4875	&	63.35	&	0.22	&	0.78	&	(0.93,	0,	0.37)	&	(0.55,	0.34,	-0.87)	\\
{\it Peanut}		&	0.61875	&	0.29375	&	0.475	&	67.48	&	0.33	&	0.67	&	(0.62,	0,	0.79)	&	(0.59,	0.48,	-0.54)	\\
{\it Black Pepper}		&	0.20625	&	0.4125	&	0.375	&	57	&	0.24	&	0.76	&	(0.89,	0,	0.45)	&	(0.37,	0.69,	-0.64)	\\
{\it Garlic}		&	0.125	&	0.7875	&	0.525	&	57.34	&	0.19	&	0.81	&	(0.94,	0,	0.35)	&	(0.47,	0.32,	-0.89)	\\
{\it Yam}		&	0.375	&	0.65625	&	0.5875	&	61.08	&	0.31	&	0.69	&	(0.61,	0,	0.79)	&	(0.7,	0.29,	-0.59)	\\
{\it Elderberry}		&	0.50625	&	0.39375	&	0.45	&	70	&	0.38	&	0.62	&	(0.7,	0,	0.71)	&	(0.65,	0.45,	-0.63)	\\
{\it Almond}		&	0.7625	&	0.29375	&	0.475	&	77.45	&	0.36	&	0.64	&	(0.87,	0,	0.49)	&	(0.59,	0.27,	-0.84)	\\
{\it Lentils}		&	0.1125	&	0.6625	&	0.375	&	72.62	&	0.24	&	0.76	&	(0.94,	0,	0.34)	&	(0.38,	0.5,	-0.81)	\\
\hline
\end{tabular}
\normalsize
\end{center}
{\rm Table 8a.} Representation of $A$, $B$ and `$A \ {\rm and} \ B$' in the case of the concepts {\it Fruits} and {\it Vegetables}. Note that the angles are expressed in degrees.
\end{table}

\begin{table}[scale=0.2]
\begin{center}
\scriptsize
\begin{tabular}{|cccccccll|}
\hline
\multicolumn{9}{|l|}{\it $A$=Fruits, $B$=Vegetables} \\
\hline					
{\it Exemplar}		&	$\mu_{x}(A)$	&	$\mu_{x}({\rm not} \ B)$	&	$\mu_{x}(A \ {\rm and} \ {\rm not} \ B)$ &	$\theta_{AB'}(x)$	&	$m_{AB'}(x)^2$	&	$n_{AB'}(x)^2$ & $|A_{AB'}(x)\rangle$ & $e^{-i \theta_{AB'}(x)}|{\rm not} \ B_{AB'}(x)\rangle$	\\
\hline
{\it Apple}		&	1	&	0.81875	&	0.8875	&	8.65	&	0.24	&	0.76	&	(1,	0,	0)	&	(0,	0.9,	-0.43)	\\
{\it Parsley}		&	0.01875	&	0.25	&	0.1	&	100.14	&	0.19	&	0.81	&	(0.99,	0,	0.14)	&	(0.07,	0.86,	-0.5)	\\
{\it Olive}		&	0.53125	&	0.44375	&	0.3375	&	88	&	0.62	&	0.38	&	(0.68,	0,	0.73)	&	(0.71,	0.23,	-0.67)	\\
{\it Chili Pepper}		&	0.1875	&	0.35	&	0.2	&	85.57	&	0.4	&	0.6	&	(0.9,	0,	0.43)	&	(0.28,	0.75,	-0.59)	\\
{\it Broccoli}		&	0.09375	&	0.0625	&	0.0875	&	62.97	&	0.24	&	0.76	&	(0.95,	0, 0.31)	&	(0.08,	0.96,	-0.25)	\\
{\it Root Ginger}		&	0.1375	&	0.325	&	0.1375	&	96.33	&	0.43	&	0.57	&	(0.93,	0,	0.37)	&	(0.23,	0.79,	-0.57)	\\
{\it Pumpkin}		&	0.45	&	0.2625	&	0.2125	&	94.72	&	0.55	&	0.45	&	(0.74,	0,	0.67)	&	(0.46,	0.72,	-0.51)	\\
{\it Raisin}		&	0.88125	&	0.7625	&	0.75	&	95.34	&	0.42	&	0.58	&	(0.94,	0,	0.34)	&	(0.18,	0.85,	-0.49)	\\
{\it Acorn}		&	0.5875	&	0.64375	&	0.4875	&	89.53	&	0.55	&	0.45	&	(0.77,	0,	0.64)	&	(0.5,	0.63,	-0.6)	\\
{\it Mustard}		&	0.06875	&	0.6	&	0.225	&	102.58	&	0.26	&	0.74	&	(0.97,	0,	0.26)	&	(0.21,	0.6,	-0.77)	\\
{\it Rice}		&	0.11875	&	0.51875	&	0.225	&	92.19	&	0.34	&	0.66	&	(0.94,	0,	0.34)	&	(0.26,	0.64,	-0.72)	\\
{\it Tomato}		&	0.3375	&	0.1875	&	0.2	&	84.39	&	0.39	&	0.61	&	(0.81,	0,	0.58)	&	(0.31,	0.85,	-0.43)	\\
{\it Coconut}		&	0.925	&	0.7	&	0.6875	&	126.44	&	0.47	&	0.53	&	(0.96,	0,	0.27)	&	(0.16,	0.82,	-0.55)	\\
{\it Mushroom}		&	0.11875	&	0.38125	&	0.125	&	105.98	&	0.45	&	0.55	&	(0.94,	0,	0.34)	&	(0.23,	0.75,	-0.62)	\\
{\it Wheat}		&	0.16875	&	0.51875	&	0.2125	&	96.33	&	0.44	&	0.56	&	(0.91,	0,	0.41)	&	(0.32,	0.61,	-0.72)	\\
{\it Green Pepper}		&	0.225	&	0.40625	&	0.2375	&	84.98	&	0.42	&	0.58	&	(0.88,	0,	0.47)	&	(0.34,	0.69,	-0.64)	\\
{\it Watercress}		&	0.1375	&	0.25	&	0.1	&	100.37	&	0.48	&	0.52	&	(0.93,	0,	0.37)	&	(0.2,	0.84,	-0.5)	\\
{\it Peanut}		&	0.61875	&	0.75	&	0.55	&	95.8	&	0.55	&	0.45	&	(0.79,	0,	0.62)	&	(0.39,	0.77,	-0.5)	\\
{\it Black Pepper}		&	0.20625	&	0.6125	&	0.2125	&	103.64	&	0.57	&	0.43	&	(0.89, 0,	0.45)	&	(0.4,	0.48,	-0.78)	\\
{\it Garlic}		&	0.125	&	0.24375	&	0.1	&	98.91	&	0.45	&	0.55	&	(0.94,	0,	0.35)	&	(0.19,	0.85,	-0.49)	\\
{\it Yam}		&	0.375	&	0.43125	&	0.2375	&	94.32	&	0.64	&	0.36	&	(0.79,	0,	0.61)	&	(0.51,	0.56,	-0.66)	\\
{\it Elderberry}		&	0.50625	&	0.60625	&	0.4125	&	89.03	&	0.59	&	0.41	&	(0.71,	0,	0.7)	&	(0.62,	0.47,	-0.63)	\\
{\it Almond}		&	0.7625	&	0.71875	&	0.6125	&	99.72	&	0.57	&	0.43	&	(0.87,	0,	0.49)	&	(0.3,	0.79,	-0.53)	\\
{\it Lentils}		&	0.1125	&	0.375	&	0.1125	&	109.72	&	0.47	&	0.53	&	(0.94,	0,	0.34)	&	(0.22,	0.76,	-0.61)	\\
\hline
\end{tabular}
\normalsize
\end{center}
{\rm Table 8b.} Representation of $A$, `${\rm not} \ B$' and `$A \  {\rm and} \ {\rm not} \ B$' in the case of the concepts {\it Fruits} and {\it Vegetables}. Note that the angles are expressed in degrees.
\end{table}

\section*{Acknowledgments.} The author is greatly indebted with  Prof. Diederik Aerts for reading the manuscript and providing a number of valuable remarks and suggestions.

\section*{References}
\begin{description}


\item Aerts, D. (1986). A possible explanation for the probabilities of quantum mechanics. {\it Journal of Mathematical Physics 27}, 202--210.


\item Aerts, D. (1999). Foundations of quantum physics: A general realistic and operational approach. {\it International Journal of Theoretical Physics 38}, 289--358.

\item Aerts, D. (2009a). Quantum structure in cognition. {\it Journal of Mathematical Psychology 53}, 314--348.

\item Aerts, D. (2009b). Quantum particles as conceptual entities: A possible explanatory framework for quantum theory. {\it Foundations of Science 14}, 361--411.

\item Aerts, D., \& Aerts, S. (1995). Applications of quantum statistics in psychological studies of decision processes. {\it Foundations of Science 1}, 85-97.

\item Aerts, D., Aerts, S., Broekaert, J., \& Gabora, L. (2000). The violation of Bell inequalities in the macroworld. {\it Foundations of Physics 30}. 1387--1414.

\item Aerts, D., Broekaert, J., Gabora, L., \& Sozzo, S. (2013). Quantum structure and human thought. {Behavioral and Brain Sciences 36}, 274--276.


\item Aerts, D., \& Czachor, M. (2004). Quantum aspects of semantic analysis and symbolic artificial intelligence. {\it Journal of Physics A: Mathematical and Theoretical 37}, L123--L132.


\item Aerts, D., \& Gabora, L. (2005a). A theory of concepts and their combinations I: The structure of the sets of contexts and properties. {\it Kybernetes 34}, 167--191.

\item Aerts, D., \& Gabora, L. (2005b). A theory of concepts and their combinations II: A Hilbert space representation. {\it Kybernetes 34}, 192--221.

\item Aerts, D., Gabora, L., \& S. Sozzo, S. (2013). Concepts and their dynamics: A quantum--theoretic modeling of human thought. {\it Topics in Cognitive Science 5}, 737--772.

\item Aerts, D., \& Sozzo, S. (2011). Quantum structure in cognition. Why and how concepts are entangled. {\it Quantum Interaction. Lecture Notes in Computer Science 7052}, 116--127.

\item Aerts, D., \& Sozzo, S. (2013). Quantum entanglement in conceptual combinations. {\it International Journal of Theoretical Physics}. DOI 10.1007/s10773-013-1946-z (in print).

\item Aerts, D., Sozzo, S., \& Tapia, J. (2014). Identifying quantum structures in the Ellsberg paradox. {\it International Journal of Theoretical Physics}. DOI DOI 10.1007/s10773-014-2086-9 (in print).

\item Aerts, D., Sozzo, S., \& Veloz, T. (2014). Negation of natural concepts and the foundations of human reasoning (in preparation).


\item Alxatib, S., \& Pelletier, J. (2011). On the psychology of truth gaps. In Nouwen, R., van Rooij, R., Sauerland, U., \& Schmitz, H.-C. (Eds.), {\it Vagueness in Communication} (pp. 13--36). Berlin, Heidelberg: Springer-Verlag.


\item Bonini, N., Osherson, D., Viale, R., \& Williamson, T. (1999). On the psychology of vague predicates. Mind and Language, 14, 377--393.


\item Busemeyer, J. R., \& Bruza, P. D. (2012). {\it Quantum Models of Cognition and Decision}. Cambridge: Cambridge University Press.


\item Busemeyer, J. R., Pothos, E. M., Franco, R., \& Trueblood, J. S. (2011). A quantum theoretical explanation for probability judgment errors. {\it Psychological Review 118}, 193--218.

\item Dirac, P. A. M. (1958). {\it Quantum mechanics}, 4th ed. London: Oxford University Press.

\item Ellsberg, D.(1961). Risk, ambiguity, and the Savage axioms. {\it Quarterly Journal of Economics 75}. 643-–669.

\item Fodor, J. (1994) Concepts: A potboiler. {\it Cognition 50}, 95--113.


\item Hampton, J. A. (1988a). Overextension of conjunctive concepts: Evidence for a unitary model for concept typicality and class inclusion. {\it Journal of Experimental Psychology: Learning, Memory, and Cognition 14}, 12--32.

\item Hampton, J. A. (1988b). Disjunction of natural concepts. {\it Memory \& Cognition 16}, 579--591.

\item Hampton, J. A. (1997). Conceptual combination: Conjunction and negation of natural concepts. {\it Memory \& Cognition 25}, 888--909.

\item Haven, E., \& Khrennikov, A. Y. (2013). {\it Quantum Social Science}. Cambridge: Cambridge University Press.


\item Kamp, H., \& Partee, B. (1995). Prototype theory and compositionality. {\it Cognition 57}, 129--191.

\item Komatsu, L. K. (1992). Recent views on conceptual structure. {\it Psychological Bulletin 112}, 500--526.

\item Khrennikov, A. Y. (2010). {\it Ubiquitous Quantum Structure}. Berlin: Springer.

\item Kolmogorov, A. N. (1933). {\it Grundbegriffe der Wahrscheinlichkeitrechnung}, Ergebnisse Der Mathematik; translated as {\it Foundations of Probability}. New York: Chelsea Publishing Company, 1950.



\item Machina, M. J. (2009). Risk, ambiguity, and the dark–dependence axioms. {\it American Economical Review 99}. 385–-392.

\item Minsky, M. (1975). A framework for representing knowledge. In P. H. Winston (Ed.) {\it The Psychology of Computer Vision} (211--277). New York: McGraw-Hill.

\item Murphy, G. L., \& Medin, D. L. (1985). The role of theories in conceptual coherence. {\it Psychological Review 92}, 289-ֳ16.

\item Nosofsky, R. (1988). Exemplar-based accounts of relations between classification, recognition, and typicality. {\it Journal of Experimental Psychology: Learning, Memory, and Cognition 14}, 700֭708.

\item Nosofsky, R. (1992). Exemplars, prototypes, and similarity rules. In Healy, A., Kosslyn, S., \& Shiffrin, R. (Eds.), {\it From learning theory to connectionist theory: Essays in honor of William K. Estes}. Hillsdale NJ: Erlbaum.

\item Osherson, D., \& Smith, E. (1981). On the adequacy of prototype theory as a theory of concepts. {\it Cognition 9}, 35--58.

\item Osherson, D. N., Smith,  E. (1982). Gradedness and Conceptual Combination. {\it Cognition 12}, 299--318.

\item Osherson, D. N., \& Smith, E. (1997). On typicality and vagueness. {\it Cognition 64}, 189--206.


\item Pitowsky, I. (1989). {\it Quantum Probability, Quantum Logic}. Lecture Notes in Physics vol. {\bf 321}.  Berlin: Springer.

\item Pothos, E. M., \& Busemeyer, J. R. (2009). A quantum probability explanation for violations of `rational' decision theory. {\it Proceedings of the Royal Society B 276}, 2171--2178.

\item Pothos, E. M., \& Busemeyer, J. R. (2013). Can quantum probability provide a new direction for cognitive modeling? {\it Behavioral and Brain Sciences 36}. 255--274.


\item Rips, L. J. (1995). The current status of research on concept combination. {\it Mind and Language 10}, 72--104.

\item Rosch, E. (1973). Natural categories, {\it Cognitive Psychology 4}, 328--350.

\item Rosch, E. (1978). Principles of categorization. In Rosch, E. \& Lloyd, B. (Eds.), {\it Cognition and categorization}. Hillsdale, NJ: Lawrence Erlbaum, pp. 133--179.

\item Rosch, E. (1983). Prototype classification and logical classification: The two systems. In Scholnick, E. K. (Ed.), {\it New trends in conceptual representation: Challenges to Piagetӳ theory?}. New Jersey: Lawrence Erlbaum, pp. 133--159.

\item Rumelhart, D. E., \& Norman, D. A. (1988). Representation in memory. In Atkinson, R. C., Hernsein, R. J., Lindzey, G., \& Duncan, R. L. (Eds.), {\it StevensҠhandbook of experimental psychology}. New Jersey: John Wiley \& Sons.



\item Sozzo, S. (2014). A quantum probability explanation in Fock space for borderline contradictions. {\it Journal of Mathematical Psychology 58}, 1--12.

\item Tversky, A. \& Kahneman, D. (1983). Extension versus intuitive reasoning: The conjunction fallacy in probability judgment. {\it Psychological Review 90}, 293--315.

\item Tversky, A., \& Shafir, E. (1992). The disjunction effect in choice under uncertainty. {\it Psychological Science 3}, 305--309.

\item Van Rijsbergen, K. (2004). {\it The Geometry of Information Retrieval}, Cambridge: Cambridge University Press.

\item Wang, Z., Busemeyer, J. R., Atmanspacher, H., \& Pothos, E. (2013). The potential of quantum probability for modeling cognitive processes. {\it Topics in Cognitive Science 5} 672--688.


\item Zadeh, L. (1965). Fuzzy sets. {\it Information \& Control 8}, 338--353.

\item Zadeh, L. (1982). A note on prototype theory and fuzzy sets. {\it Cognition 12}, 291--297.

\end{description}

\end{document}